\documentclass[12pt]{article}

\usepackage{arxiv}

\usepackage[utf8]{inputenc} 
\usepackage[T1]{fontenc}    
\usepackage{hyperref}       
\usepackage{url}            
\usepackage{booktabs}       
\usepackage{amsfonts}       
\usepackage{nicefrac}       
\usepackage{microtype}      
\usepackage{lipsum}
\usepackage{graphicx}
\RequirePackage{amsmath}
\usepackage{tabularx}
\usepackage{algorithm,algpseudocode}
\usepackage{graphicx,subcaption}

\title{Generative Learning of Densities 
on Manifolds}

\author{
Dimitris G. Giovanis\thanks{Corresponding author}\\
  Dept. of Civil and Systems Engineering\\
  Johns Hopkins University\\
  Baltimore, MD, USA \\
  \texttt{dgiovan1@jhu.edu} \\
   \And
 Ellis Crabtree \\
  Dept. of Chemical and Biomolecular Engineering\\
  Johns Hopkins University\\
  Baltimore, MD, USA\\
  \texttt{ecrabtr2@jhu.edu} \\
  \And
Roger G. Ghanem \\
Dept. of Civil and Environmental Engineering  \\
Dept. of Aerospace and Mechanical Engineering \\
University of Southern California\\
Los Angeles, CA, USA \\
  \texttt{ghanem@usc.edu} \\
   \AND
 Ioannis G. Kevrekidis\thanks{Corresponding author} \\
  Dept. of Chemical and Biomolecular Engineering\\
  Dept. of Applied Mathematics and Statistics \\
  Johns Hopkins University\\
  Baltimore, MD, USA\\
  \texttt{yannisk@jhu.edu}
}

\begin{document}
\maketitle
\begin{abstract}
A generative modeling framework is proposed that combines  diffusion models and manifold learning to efficiently sample data densities on manifolds. The approach utilizes Diffusion Maps to uncover possible low-dimensional underlying (latent) spaces in the high-dimensional data (ambient) space. Two approaches for sampling from the latent data density are described. The first is a score-based diffusion model, which is trained to map a standard normal distribution to the latent data distribution using a neural network. The second one involves solving an  Itô  stochastic differential equation in the latent space.  Additional realizations of the data are generated by lifting the samples back to the ambient space using {\em Double Diffusion Maps}, a recently introduced technique typically employed in studying dynamical system reduction; here the focus lies in sampling densities rather than system dynamics. The proposed approaches enable sampling high dimensional data densities restricted to low-dimensional, \textit{a priori unknown} manifolds. The efficacy of the proposed framework is demonstrated through a benchmark problem and a material with multiscale structure. 
\end{abstract}


\section{Introduction}
\noindent
Generative modeling has emerged as a transformative approach, enabling simulation and providing unprecedented opportunities for the study of complex systems. By learning the data densities, generative models (GM) can produce new samples that reflect the statistical properties of observed phenomena. Beyond data generation, GMs can also improve our understanding of the mechanisms driving these phenomena and help update system digital twins. This capability has facilitated scientific discovery in fields such as molecular biology \cite{jumper2021highly}, neuroscience \cite{wang2023applications},  physics \cite{de2017learning}, materials \cite{dan2020generative}, and climate science \cite{besombes2021producing}, to name some.  As the field of generative modeling continues to evolve, its integration with domain-specific knowledge promises to unlock new scientific discoveries.

Modern GMs are fundamentally built on the concept of mapping one probability distribution to another, enabling the transformation of samples from the source distribution into those of the target distribution. These approaches leverage the remarkable expressive capabilities and adaptability of deep neural networks (DNN) to identify the underlying generative SDEs. 
Generative models are typically categorized into three main types: likelihood-based approaches, such as Variational Autoencoders (VAE) \cite{kingma2013auto, rezende2014stochastic}, autoregressive methods \cite{larochelle2011neural}, and normalizing flows \cite{papamakarios2021normalizing}; 
implicit models, such as Generative Adversarial Networks (GAN) \cite{goodfellow2020generative, mirza2014conditional};
and diffusion-based models such as Score-based GMs \cite{song2020score}. 

The latter stand out by employing Stochastic Differential Equations (SDEs) to incrementally introduce noise into data distributions, followed by a reverse process to generate new samples \cite{song2020score}. These models estimate the gradient field, or \textit{score}, of the modified data distribution, enabling high-quality data generation through numerical integration of the reverse SDE. Approaches that fall into this category include Langevin dynamics-based score matching \cite{song2019generative} and denoising diffusion probabilistic models (DDPM) \cite{ho2020denoising}, to improve sampling efficiency and likelihood computation.   Despite the success of score-based generative/diffusion models (SGM) in generating data (images \cite{song2019generative, song2020improved, ho2020denoising}, audio \cite{chen2020wavegrad, kong2020diffwave}, graphs \cite{niu2020permutation}, and shapes \cite{cai2020learning}),  their efficiency for density estimation poses unique computational and statistical challenges, as we discuss next.  

Generative models for density estimation are commonly formulated as tasks without explicit labels. To train the underlying deep neural networks in these models, various loss functions have been developed, such as adversarial loss for GANs \cite{goodfellow2014generative}, maximum likelihood loss for normalizing flows \cite{kobyzev2020normalizing}, and score matching losses for diffusion models \cite{hyvarinen2005estimation, vincent2011connection}. While these approaches have achieved significant success, training poses several challenges. For instance, GANs often suffer from issues like mode collapse, vanishing gradients, and instability. Likewise, maximum likelihood loss in normalizing flows requires efficient computation of the Jacobian determinant, imposing constraints on network architecture design.
These training challenges can be mitigated if the generator in generative models, such as the decoder in VAEs or normalizing flows, is trained with explicit guidance. An alternative framework proposed in \cite{liu2024diffusion} introduces a structured approach for training generative models for density estimation. This framework employs a score-based diffusion model (SGM) to generate reference data, which is then used to train a deep neural network with a straightforward mean-squared error (MSE) loss. The reverse-time SDE used for generating reference data relies on a training-free score estimation method based on mini-batch-based Monte Carlo simulation (MCS), enabling direct approximation of the score function at any spatial-temporal location and simplifying the process of solving the reverse-time SDE.

Another limitation of GMs involves sampling from densities in the presence of small imbalanced datasets and/or when the data have a manifold-like structure. In this work, we are interested in tackling the latter one. In such cases, the generated data may not accurately preserve the intrinsic geometry of the manifold \cite{aggarwal2001surprising} and specific regions of the data distribution will be underrepresented. To address this issue, nonlinear dimension reduction (or manifold learning) can be utilized to restrict the generated samples to lie on the manifold. Manifold learning is based on the assumption that high-dimensional data typically lie on a lower-dimensional manifold embedded within the high-dimensional space \cite{fefferman2016testing}. Examples of manifold learning methods include Diffusion Maps  \cite{coifman2006diffusion}, Isomap \cite{balasubramanian2002isomap}, and local linear embedding (LLE) \cite{roweis2000nonlinear}. Over the last several years, manifold learning has been utilized for sampling densities. Probabilistic Learning on the Manifold (PLoM), introduced by Soize and Ghanem in a series of papers \cite{soize2016data, soize2019entropy, soize2021probabilistic, soize2022probabilistic}, utilizes Diffusion Maps to generate additional realizations of random vectors whose probability distribution is concentrated on an unknown low-dimensional manifold embedded in a higher-dimensional space. The method is based on solving a reduced-order It\^o SDE. 

In this work, we present a framework for sampling from data densities that are restricted on low-dimensional manifolds that combines Double Diffusion Maps \cite{evangelou2023double} with: (1) score-based SGM \cite{liu2024diffusion} and (2) an It\^o SDE \cite{soize2016data}. Double diffusion Maps were introduced and initially used for building reduced dynamical models; the reduced latent space is discovered using Diffusion Maps. A second round of Diffusion Maps on those latent coordinates allows the approximation of the reduced dynamical models. This second round enables \textit{lifting} the latent space
coordinates back to the full ambient space through the \textit{Nystr\"om Extension} \cite{coifman2006geometric}. The main contribution of this work is that we utilize diffusion models to sample from the densities in the latent space discovered with Diffusion Maps. As a result, the complexity of the problem is reduced. Double Diffusion Maps then allow lifting the generated data (with any of our two methods) back to the original space, satisfying the manifold-based structure.

The remainder of this paper is structured as follows. Section~\ref{sec:score_diffusion} provides an overview of diffusion models and their mathematical foundations.  The manifold learning tools used in this work are discussed in~\ref{sec:ddmaps}, \ref{sec:GH} and \ref{sec:plom}.
In Section \ref{sec:problem} we discuss the motivation for this work followed by the proposed framework for sampling densities in Section~\ref{sec:method}. Section~\ref{sec:examples} presents numerical results to demonstrate the efficiency of our approach.  Finally, Section~\ref{sec:conclusions}
provides insight and future directions.

\section{Score-based diffusion models}
\label{sec:score_diffusion}

\noindent
We start with a set of independent and identically distributed samples $\mathcal{X}=\{\textbf{x}_1, \textbf{x}_2, \ldots, \textbf{x}_N\}$, where $\textbf{x}_i \in \mathbb{R}^d$, are observed from the probability density function (pdf) $f_\textbf{x}(x_1, x_2, \ldots, x_d)$. Our goal is to generate samples from this density.
Consider the complete filtered probability space $\{\Omega, \mathcal{F}, \mathbb{P}, \{\mathcal{F}_t\}_{0\leq t\leq T}\}$, where $\Omega$ is the sample space, $\mathcal{F}$ is the $\sigma$-algebra, $\mathbb{P}$ is the probability measure, and 
$\{\mathcal{F}_t\}_{0\leq t\leq T}$ is the natural filtration of an $d$-dimensional Wiener process (Brownian motion) $W_t$, i.e., an increasing sequence of sigma-algebras, representing the information available over time. As  $t$ grows,  $\mathcal{F}_t$ contains more events, representing our knowledge accumulation. 

 We can define a continuous-time diffusion process \( \{x(t)\}_{t=0}^T \) in the probability space  $\{\Omega, \mathcal{F}, \mathbb{P}, \{\mathcal{F}_t\}_{0\leq t\leq T}\}$, such that $ x(0) \sim f_\textbf{x}(\cdot)$ is the target (data) distribution, and \( x(T)  \sim f_\textbf{y}(\cdot) \), where $f_\textbf{y}(\cdot)\in \mathbb{R}^d$ is some ``prior'' distribution, that is easy to sample from. This process can be modeled by forward and reverse-time Itô stochastic differential equations (SDEs), defined in a standard temporal domain $t \in [0, 1]$. The forward SDE is defined as:
\begin{equation}\label{eq:forwardSDE}
\text{d}x_t = g(x, t) \, \text{d}t + \sigma(t) \, \text{d}W_t, \quad x_0 = x(0), \quad x_1 = x(1),
\end{equation}
where  \( g(\cdot, t): \mathbb{R}^d \to \mathbb{R}^d \) is the drift coefficient, \( \sigma(\cdot): \mathbb{R} \to \mathbb{R} \) is the diffusion coefficient, $x_0 = x(0)$ is the initial state, and $x_1 = x(1)$ is the terminal state of the diffusion process.  For simplicity, \( \sigma \) is assumed to be a scalar function independent of \( x \), but this can be generalized.  
The forward SDE is utilized to map  $f_\textbf{x}(\cdot) \rightarrow f_t(x)  \equiv f(x_t) \equiv f(x, t)\rightarrow f_\textbf{y}(\cdot)$, where $f_t$ are intermediate  (at time $t$) densities.  A number of choices for the drift and diffusion coefficients ensure that the terminal state of the diffusion process follows  $f_\textbf{y}(\cdot) \sim \mathcal{N}(\textbf{0}, \textbf{I}_d)$ \cite{ho2020denoising, song2020score}. 

According to \cite{ho2020denoising}, the reverse of a diffusion process is itself a diffusion process with the following reverse-time SDE:
\begin{equation}
\text{d}x = \left[ g(x, t) - \sigma^2(t) \nabla_{x_t} \log f_t(x) \right] \, \text{d}t + \sigma(t) \, \text{d}\overleftarrow{W}_t, \quad x_0 = x(0), \quad x_1 = x(1)
\label{eq:A3}
\end{equation}
where \( \overleftarrow{W}_t \) is a standard Wiener process running backward in time from \( t \) to \( 0 \), and \( \text{d}t \) is a negative infinitesimal time step. If the score \( S(x_t, t)=\nabla_x \log f_t(x) \) is known for all \( t \in [0, 1] \), we can derive the reverse diffusion process from Eq.(\ref{eq:forwardSDE}) and use it to sample from the density \( f_0=f_{\textbf{x}}(\cdot) \). The evolution of the probability density $f_t$ is described by the Fokker-Planck equation:

\begin{equation}
\frac{\partial f(x, t)}{\partial t} = -\nabla \cdot \big [g(x, t) - \sigma^2(t)\nabla_{x_t} \log f_t(x) -\frac{\sigma^2(t)}{2} \nabla f(x, t) \big ]
\end{equation}
where $S(x_t, t) = \nabla_{x_t} \log f(x_t)$ is the gradient of the log-probability density function \( f(x_t) \) with respect to \( x_t \), and is  called the \textit{score function}. To estimate $S(x_t, t)$, various approaches exist that  include score matching \cite{hyvarinen2005estimation}, denoising score matching \cite{vincent2011connection}, and sliced score matching \cite{song2020sliced}. Recent progress in diffusion models has focused on using neural networks to approximate the score function. Unlike normalizing flow models \cite{papamakarios2021normalizing}, a key limitation of learning the score function is that sampling requires solving the reverse-time SDE. To overcome this limitation, a training-free score-based diffusion model for data labeling was introduced in \cite{liu2024diffusion},  facilitating supervised learning of the generative model. In this work we utilize the approaches, discussed next.

\subsection{Estimation of the score function}

\subsubsection{Monte Carlo-based (MCS) estimation}
\label{sec:diffusion_mcs}
\noindent
Several acceptable forms exist for the drift and diffusion of Eq.~(\ref{eq:forwardSDE}). Among these, specific parameterizations ensure that the terminal state is Gaussian \cite{song2020score, ho2020denoising, lu2022dpm}. Following these formulations, we adopt the following definitions \cite{liu2024diffusion}:

\begin{equation}
    g(x, t) = b(t) \cdot x_t = \frac{\text{d} \log \alpha_t}{\text{d}t} \cdot x_t, \quad 
    \sigma^2(t) = \frac{\text{d} \beta^2_t}{\text{d}t} - 2 \frac{\text{d} \log \alpha_t}{\text{d}t} \beta^2_t,
\end{equation}
where the functions
\[
\alpha_t = 1 - t, \quad \beta_t^2 = t, \quad t \in [0, 1],
\]
are chosen to guarantee the Gaussianity of the terminal state.  Since the forward SDE remains linear under these parameterizations, the conditional pdf, \( f(x_t \mid x_0) \), for a fixed \( x_0 \), is Gaussian:
\[
f(x_t \mid x_0) \sim \mathcal{N}(\mu_t(x_0), \Sigma_t),
\]
where \( \mu_t(x_0) \) and \( \Sigma_t \) are the mean and covariance derived explicitly from the dynamics of the SDE

\begin{equation}\label{forward_SDE}
    f(x_t|x_0) = f_t = \mathcal{N}(\alpha_t  x_0, \beta_t\textbf{I}_d).
\end{equation}
Equation (\ref{forward_SDE}) can be derived from the solution of the forward SDE, given by:
\begin{equation}
x_t = \alpha_t x_0 + \int_0^t \sigma(s) \, dW_s,
\end{equation}
where \( \alpha_t = \exp \left( \int_0^t b(s) \, ds \right) \). Using the definitions of \( \alpha_t = 1 - t \) and \( \beta_t^2 = t \), one can prove that \( x_t \) has mean \( \mathbb{E}[x_t | x_0] = \alpha_t x_0 \) and variance \( \text{Var}(x_t | x_0) = \beta_t^2 \textbf{I}_d \). Thus, \( f(x_t | x_0) \sim \mathcal{N}(\alpha_t x_0, \beta^2_t \textbf{I}_d) \). For $t=1$, this leads to  \( f_1=f(x_1 | x_0) \sim \mathcal{N}(\textbf{0}, \textbf{I}_d) \), since $\alpha_1=0$ and $\beta_1=1$.

The law of total probability connects the joint probability density $f(x_t | x_0)$  with the marginal densities $f(x_t)$ as:

\begin{equation}
f(x_t) = \int_{\mathbb{R}^d} f(x_t,  x_0) \text{d}x_0 = \int_{\mathbb{R}^d} f(x_t | x_0) f(x_0) \, \text{d}x_0
\end{equation}
where  \( f(x_t | x_0) \) represents the conditional probability density of \( x_t \) given \( x_0 \), and \( f(x_0) \) is the probability density of \( x_0 \). Applying the chain rule we can write the score function as:
\begin{eqnarray}
S(x_t, t) &=& \nabla_{x_t} \log \left( \int_{\mathbb{R}^d} f(x_t | x_0) f(x_0) \, \text{d}x_0 \right) \\ \nonumber
&=& \frac{ \nabla_{x_t}  \int_{\mathbb{R}^d} f(x_t | x_0) f(x_0) \, \text{d}x_0}{\int_{\mathbb{R}^d} f(x_t | x_0) f(x_0) \text{d}x_0}\\ 
&=&\frac{ \int_{\mathbb{R}^d} \nabla_{x_t}  f(x_t | x_0) f(x_0) \, \text{d}x_0}{\int_{\mathbb{R}^d} f(x_t | x_0) f(x_0) \text{d}x_0}.
\label{eq:A2}
\end{eqnarray}
Since the conditional distribution \( f(x_t | x_0)\) is Gaussian, i.e.,  \( f(x_t | x_0) \sim \mathcal{N}(\alpha_t Z_0, \beta_t^2 I) \), the gradient \( \nabla_{x_t} f(x_t | x_0) \) with respect to \( x_t \) is\footnote{The proof is provided in \ref{sec:proof}.}:
\begin{equation}
\nabla_{x_t} f(x_t | x_0) = - \frac{x_t - \alpha_t x_0}{\beta_t^2} f(x_t | x_0) 
\label{eq:A1}
\end{equation} 
Substituting Eq.~(\ref{eq:A1}) into  Eq.~(\ref{eq:A2}), we obtain:
\begin{eqnarray}
S(x_t, t) &=& \frac{1}{\int_{\mathbb{R}^d} f(x_t | x_0) f(x_0) \, \text{d}x_0} \int_{\mathbb{R}^d} \frac{x_t - \alpha_t x_0}{\beta_t^2} f(x_t | x_0) f(x_0) \, \text{d}x_0 \\ \nonumber &=& \int_{\mathbb{R}^d}\frac{x_t - \alpha_t x_0}{\beta_t^2} w_t(x_t | x_0) f(x_0) \, \text{d}x_0,
\end{eqnarray}
where the weight function \( w_t(x_t, x_0) \) is defined as
\begin{equation}
w_t(x_t, x_0) = \frac{f(x_t | x_0)}{\int_{\mathbb{R}^d} f(x_t | x_0) f(x_0). \, \text{d}x_0}
\end{equation}
The weight function \footnote{The purpose of  weighting  is to normalize the terms in the summation to ensure they form a valid probability distribution i.e., $\sum_{n=1}^N w_t(x_t, x_j^n) = 1$, and thus making \( \tilde{S}(x_t, t) \) an unbiased estimator of \( S(x_t, t) \)} can be approximated using mini-batch MCS
based on the available samples $\{\textbf{x}_1, \textbf{x}_2, \ldots, \textbf{x}_N\}$ as:
\begin{equation}
w_t(x_t, x_j^n) \approx \overline{w}(x_t, x_j^n) = \frac{f(x_t | x_j^n)}{\sum_{m=1}^N f(x_t | x_j^m)}
\end{equation}
where $\{x_{j}^{n}\}_{n=1}^{N_m}$ represents a mini-batch of the data, with $N_m \leq N$. Finally, the score function, at each time step $t$, is approximated by:
\begin{equation}
\tilde{S}(x_t, t) = \sum_{n=1}^{N_m} \frac{x_t - \alpha_t x_j^n}{\beta_t^2} \overline{w}_t(x_t, x_j^n).
\label{eq:SGM}
\end{equation}

Overall, starting with samples \( \{\textbf{y}_m\}_{m=1}^M \) drawn from \( \mathcal{N}(0, \textbf{I}_d) \), labeled data  \( \{\textbf{x}_m\}_{m=1}^M \)   are generated  by solving the reverse-time SDE from \( t=1 \) to \( t=0 \). The dataset \( D_{\text{train}} = \{(\textbf{y}_m, \textbf{x}_m)\}_{m=1}^M \) is utilized to train, in a supervised manner, a feed-forward neural network, based on minimizing the mean squared error (MSE), to learn the mapping \( \mathcal{N}(0, \textbf{I}_d) \rightarrow f_{\textbf{x}}(\cdot) \).

Since we are interested in generating  labeled data for training the neural network, we would like the mapping \( \mathcal{N}(0, \textbf{I}_d) \rightarrow f_{\textbf{x}}(\cdot) \) to be ``smooth''. Given that an ordinary differential equation (ODE) can create a much smoother functional relationship between the initial state and the terminal state, compared to the SDE,  we consider the Liouville equation that describes the evolution of probability density under deterministic flow:
\begin{equation}
\frac{\partial f(x, t)}{\partial t} = -\nabla \cdot \big [ h(x, t) f(x, t)\big ]
\end{equation}
where $h(x, t)=h(x, t)= b(t)\cdot x - \sigma^2(t)S(x, t) -\frac{\sigma^2(t)}{2} \nabla \log f(x, t) $. Instead of the reverse-time SDE of Eq.~(\ref{eq:A3}), we can solve the following ODE that approximates the reverse-time SDE:
\begin{equation}
\text{d}x_t = \left( g(x, t) - \frac{1}{2} \sigma^2(t) S(x_t, t) \right) \, \text{d}t
\end{equation}
After the neural network is trained, this method does not require solving the reverse-time ODE (or SDE) to generate samples of $f_{\textbf{x}}(\cdot)$, which makes the approach computationally  efficient.

\subsubsection{Score-Matching estimation}
\noindent
The more standard literature approach to estimate the score function involves using a neural network to approximate $\nabla_x \log \hat{f}_t(x)$ (where $\hat{f}_t(x)$ refers to an approximation of the distribution ${f}_t(x)$). This can be done by using Score-Matching \cite{song2020improved, song2020score}, resulting in the following loss for the neural network approximation of the reverse SDE drift:
$${L}(\theta) = \mathbb{E}_{t \sim \mathcal{U}(0,T)}\mathbb{E}_{x(0) \sim p(x(0))}\mathbb{E}_{x(t) \sim f(x(t) | x(0))}[\lambda (t) \| \nabla \log \hat{f}_t(x) - s_\theta(x(t), t)\|^2_2],$$
where $\mathcal{U}(0,T)$ is a uniform distribution over the time interval of [0,T], $\lambda (t)$ is a positive weighting function, and $x(0)$ denotes $x(t)$ at time $t=0$. Note that this score-matching loss is comparable to a more computationally efficient \emph{denoising score matching objective} \cite{vincent2011connection}, which can be used as a substitute loss function.  The final result is a neural-net based model that utilizes this discretized SDE to generate data consistent with target conditional distributions. All SDE parameters for the SGM model used in this work were chosen using guidance from Karras et. al. \cite{karrasdiffparams}.

\section{Problem statement: sampling densities with low-dimensional intrinsic structure}
\label{sec:problem}
\begin{figure}[ht!]
    \centering 
        \begin{subfigure}[b]{0.45\textwidth}
        \centering
        \includegraphics[width=\textwidth, trim={1cm 3cm 1cm 2cm}, clip]{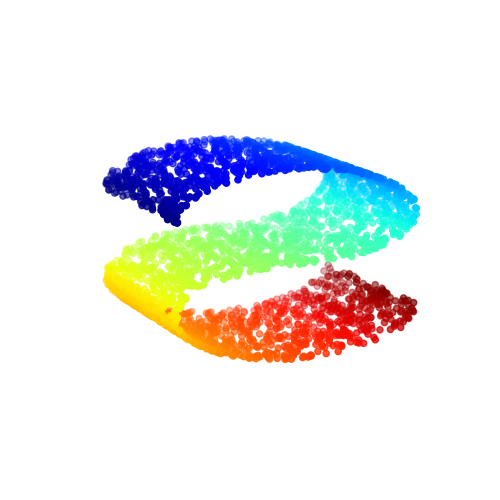}
        \caption{}
        \label{fig:sub12}
    \end{subfigure}
    \hfill
    \begin{subfigure}[b]{0.49\textwidth}
        \centering
        \includegraphics[width=\textwidth, trim={1cm 3cm 1cm 2cm}, clip]{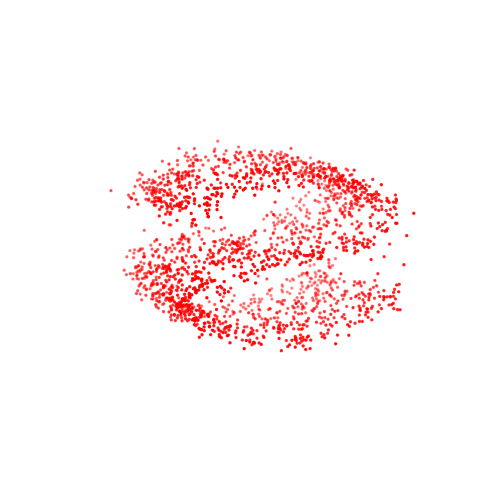}
        \caption{}
        \label{fig:sub22}
    \end{subfigure}
    \caption{(a) Three-dimensional S-shaped data  with 10,000 points. (b) 5,000 samples generated using the MCS-based SGM.}
    \label{fig:S_curve_data}
\end{figure}

\noindent
While diffusion models have proven to be very efficient in sampling from complex data densities, challenges occur when the density has an underlying lower-dimensional structure. Let us consider the example of three-dimensional S-shaped surface. The data structure  is  represented by points on a two-dimensional manifold embedded in three-dimensional space. For an S-surface parameterized by 
\( t\in [-\pi,\pi]\) and  $h$, a uniform random variable in [-1,1], the coordinates \( (x, y, z) \) of a point on the surface can be defined as:

\[
x = \sin\left(\frac{\pi}{2} t\right),
\quad y = h,
\quad z = \operatorname{sign}(t) \left(1 - \cos\left(\frac{\pi}{2} t\right)\right),
\]
where \( t \) controls the ``arclength'' position along the S-backbone 
and \( h \) provides the ``width'' of the S-shaped ribbon. Figure \ref{fig:sub12} depicts 10,000 points $\textbf{x} =\{x, y, x\} \sim f_{\textbf{x}}(\cdot)$, colored according to the values of $t$.

\begin{figure}[!htb]
        \centering
        \vspace{-1em} 
        \includegraphics[width=0.8\textwidth]{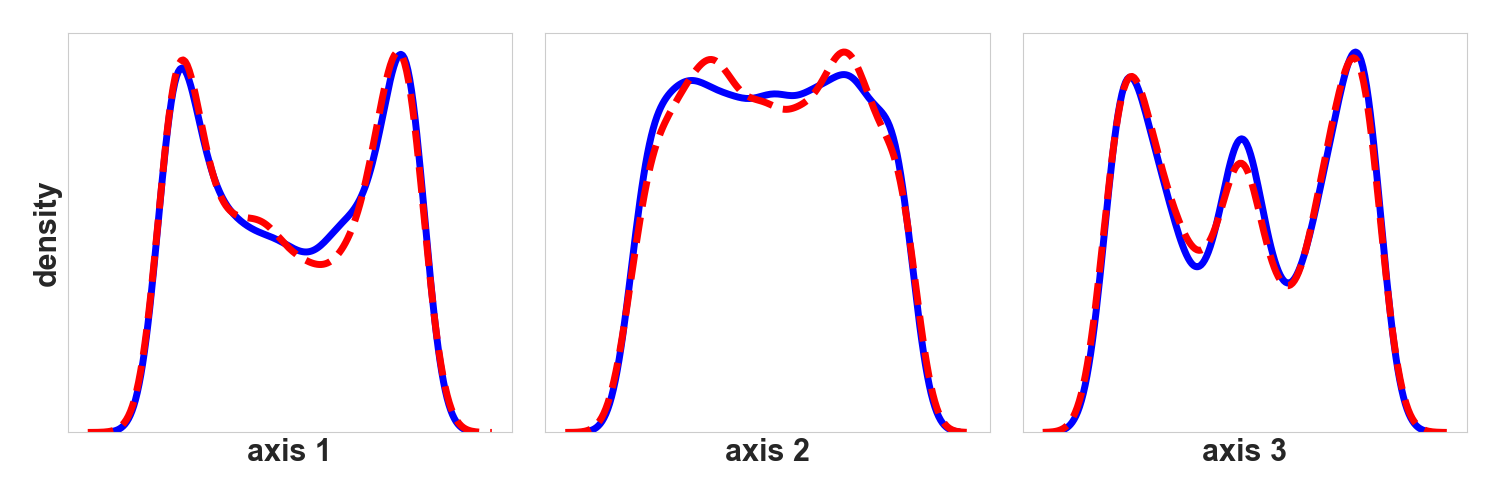}
    \caption{Comparison of the three SGM  marginal densities (red dashed lines) against the ground truth densities (blue solid lines)}
    \label{fig:S_curve_data_densities}
\end{figure}

Using the MCS-based SGM, we generate 5,000 points from $f_{\textbf{x}}(\cdot)$ (see Fig.~\ref{fig:sub22}). The marginal densities of the ground truth data (blue lines) and of the sampled data (red  dashed lines) are depicted in Fig.~\ref{fig:S_curve_data_densities}. We  notice that there is very good visual agreement between the true and the generated densities.

However, a closer inspection of the generated data in Fig.~\ref{fig:sub12} reveals that the generated points \textit{extend beyond the S-shaped manifold}. This behavior is expected, as the generative model utilizes MCS to sample from the data density without enforcing constraints that restrict the points to respect the two-dimensional geometry of the data. This behavior is further illustrated in Fig.~\ref{fig:sub22}, which presents three two-dimensional projections of the data; we can see that the red points extend beyond the S-shaped manifold, exhibiting a thickness (variance) around it.

\begin{figure}[ht!]
    \centering 
\includegraphics[width=\textwidth]{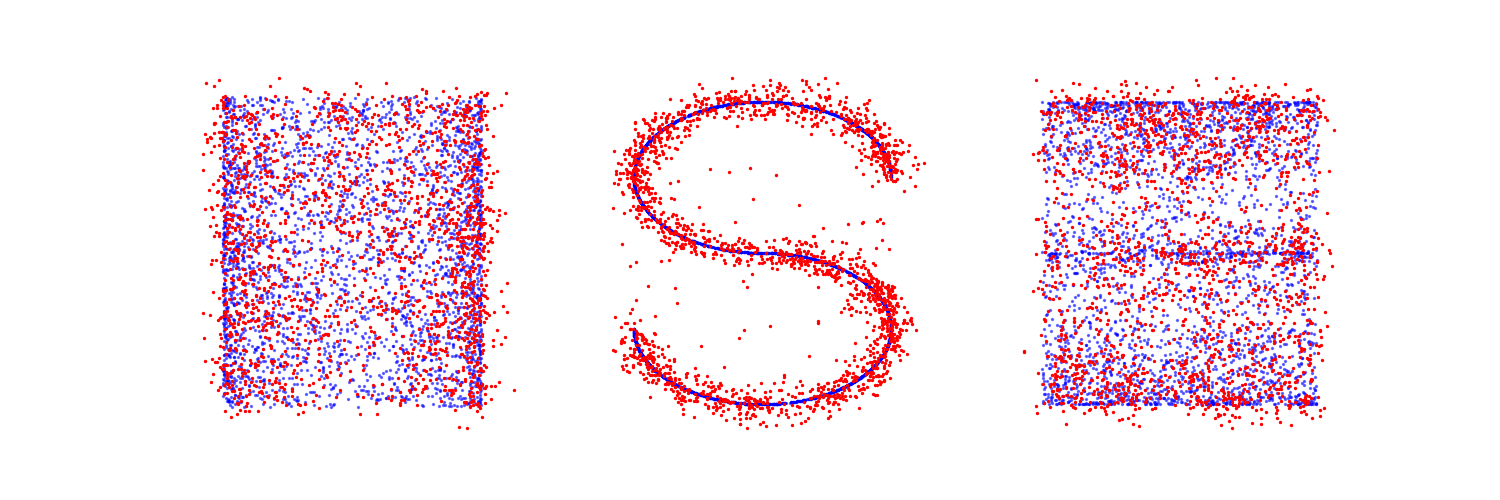}
    \caption{Two-dimensional projections of the original (blue) and the sampled (red) data.}
    \label{fig:SGM-S_curve-SGM_sections}
\end{figure}

Given this limitation of the supervised generative model, the objective of this work is to extend its capability to sample from densities that have an underlying low-dimensional structure, in a way that the generated data  \textit{respect the structure of the manifold}. To achieve this, we propose to integrate the recently proposed manifold learning with Double Diffusion Maps, method into the framework of generative modeling.

\section{Proposed Framework}
\label{sec:method}

\noindent
In the core of the proposed method lies manifold learning using Diffusion Maps. The Diffusion Maps algorithm \cite{coifman2006diffusion} can be used in two primary ways: (1) for dimensionality reduction of a finite dataset, $\mathcal{X} = \{ \textbf{x}_i \}_{i=1}^N$ where $\textbf{x}_i \in \mathbb{R}^d$, sampled from a manifold $\mathcal{M} \subset \mathbb{R}^d$ (see \ref{sec:dmaps_red}); and/or (2) to construct a basis of functions called Geometric Harmonics (GH) that extends functions defined on $\mathcal{X}$, i.e., $h(\mathcal{X}) : \mathcal{X} \rightarrow \mathbb{R}$ (see \ref{sec:GH} )\cite{lafon2004diffusion, coifman2006geometric}. Here, we utilize the recently proposed Double Diffusion Maps \cite{evangelou2023double}, which applies Diffusion Maps successively, as detailed in \ref{sec:ddmaps}. The basic steps of the approach are:
\begin{itemize}
    \item Using Diffusion Maps to identify from the data latent variables $\boldsymbol{\Phi}$, supporting the data density $f_{\boldsymbol{\Phi}}(\cdot)$;
    \item Learn the density $f_{\boldsymbol{\Phi}}(\cdot)$ and generate additional data from it;
    \item \textit{Lift} the generated data back into the ambient space using Double Diffusion Maps.
\end{itemize}
To learn the density $f_{\boldsymbol{\Phi}}(\cdot)$ we utilize: (1) the diffusion model described in Section \ref{sec:score_diffusion}.  (2) Probabilistic Learning on the Manifold (PLoM), an alternative approach to sampling from densities, proposed in 2016 by Ghanem and Soize \cite{soize2016data}. The details of PLoM are presented in \ref{sec:plom}.
Algorithms \ref{alg:SGM}, \ref{alg:nn_SGM} present the framework based on the score-based diffusion \textit{manifold} model (termed m-SGM) for the \textit{manifold} Monte Carlo (m-SGM1) and the \textit{manifold} Score-Matching (m-SGM2) cases, respectively. Algorithm \ref{alg:Ito} depicts the \textit{manifold} PLoM  model (termed m-PLoM). Both approaches consider an initial dataset constiting of a small number of points, generated from e.g., a computational model or an experiment. The  m-SGM method constructs a (supervised) machine learning model that samples from the data density on the latent space $f_{\boldsymbol{\Phi}}(\cdot)$ based on the following steps: (i) discover the intrinsic geometry of the dataset (if it exists), (ii) learn the mapping from the identified latent space back to the ambient space, and its functions, (iii) utilize MCS-based SGM to generate labeled data in the latent space by solving the reverse-time ODE, (iv) train a neural network to learn the mapping between the points in the (appropriately dimensional, reduced) standard normal space and the latent space, (v)  use the trained neural network to generate additional realizations of the data in this latent space, and (vi) map the generated points back to the ambient space. 
On the other hand, m-PLoM consists of projecting the latent variables onto the basis of the non-trivial principal components (PCA), finding their Diffusion Maps representation, and finally solving a reduced It\^o SDE to sample from $f_{\boldsymbol{\Phi}}(\cdot)$; these are then lifted back to the ambient space using Double Diffusion Maps.

\begin{algorithm}[H]
\caption{m-SGM1: Monte Carlo-based diffusion model}\label{alg:SGM}

\begin{algorithmic}[1]
\Require A dataset $\{\boldsymbol{\phi}_i\}_{i=1}^N \subset \mathbb{R}^m \sim f_{\boldsymbol{\Phi}}(\cdot)$
\State \textbf{Reverse-time ODE}:
\begin{itemize}

\item Generate $N$ points from the normal distribution, i.e.,  $\textbf{y} \sim\mathcal{N}(\textbf{0}, \textbf{I}_d) \subset \mathbb{R}^m$

\item Compute the score function $\tilde{S}(x_t, t)$ in Eq.~(\ref{eq:SGM}) and generate labeled data $\{ \boldsymbol{\phi}_i\}_{i=1}^N$

\end{itemize}

\State \textbf{Neural Network}
\begin{itemize}
\item Train a feed-forward neural network to learn the functional $\boldsymbol{\phi} = L(\textbf{y})$.
\item Draw $N^\star$ samples from $\mathcal{N}(\textbf{0}, \textbf{I}_d)$ and use the neural network to predict  $\{\boldsymbol{\tilde{\phi}}_i\}_{i=1}^{N^\star}$
\end{itemize}

\State \textbf{Lifting Operations}
\begin{itemize}
\item Use Double Diffusion Maps and Latent Harmonics to lift the predicted points to the ambient space: $\{\boldsymbol{\tilde{\phi}}_i\}_{i=1}^{N^\star} \rightarrow \{\tilde{\textbf{x}}_i\}_{i=1}^{N^\star} \sim f_{\textbf{x}}(\cdot)$
\end{itemize}

\end{algorithmic}
\end{algorithm}

\begin{algorithm}[H]
\caption{m-SGM2: Score-Matching diffusion  model}\label{alg:nn_SGM}
\begin{algorithmic}[1]
\Require A dataset $\{\boldsymbol{\phi}_i\}_{i=1}^N \subset \mathbb{R}^m \sim f_{\boldsymbol{\Phi}}(\cdot)$
\State \textbf{Approximate the score function}:
\begin{itemize}

\item Initialize a feed-forward neural network

\item Train the neural network using a score-matching loss function $L(\theta)$

\end{itemize}

\State \textbf{Generate new points}
\begin{itemize}

\item Utilize the neural-net based model to generate  $N^\star$ data consistent with target conditional distribution $f_{\boldsymbol{\Phi}}(\cdot)$
\end{itemize}

\State \textbf{Lifting Operations}
\begin{itemize}
\item Use Double Diffusion Maps and Latent Harmonics to lift the predicted points back to the ambient space: $\{\boldsymbol{\tilde{\phi}}_i\}_{i=1}^{N^\star} \rightarrow \{\tilde{\textbf{x}}_i\}_{i=1}^{N^\star} \sim f_{\textbf{x}}(\cdot)$
\end{itemize}
\end{algorithmic}
\end{algorithm}

\begin{algorithm}[!htb]
\caption{m-PLoM}\label{alg:Ito}

\begin{algorithmic}[1]
\Require A matrix version of the dataset $[\boldsymbol{\Phi}] = [\{\boldsymbol{\phi}_i\}_{i=1}^N] \subset \mathbb{R}^m \sim f_{\boldsymbol{\Phi}}(\cdot)$
\State \textbf{Principal Component Analysis}:
\begin{itemize}

\item Perform its PCA using Eq.~\ref{eq:pca} and obtain matrix $[\textbf{H}]$: $[\boldsymbol{\Phi}] =[x] + [\phi] [\mu]^{1/2} [\textbf{H}]$

\end{itemize}

\State \textbf{Nonparametric statistical estimate of the density}
\begin{itemize}
\item Estimate the density $f_{\textbf{H}}(\boldsymbol{\eta})$ using Eq.\ref{eq:pdf}.
\end{itemize}

\State \textbf{Nonlinear I\^to stochastic differential equation (ISDE)}
\begin{itemize}
\item Find Diffusion Map basis $[g]$ in this PCA space, and express the matrix $[\textbf{H}]=[\textbf{Z}][g]^{\top}$

\item Solve a reduced-order It\^o SDE to generate $N^\star$ additional realizations of matrix $[\textbf{H}]$.

\item Inverse Eq.~\ref{eq:pca} to obtain $[\tilde{\boldsymbol{\Phi}}] = [\{\boldsymbol{\tilde{\phi}}_i\}_{i=1}^{N^\star}]$
\end{itemize}

\State \textbf{Lifting Operations}
\begin{itemize}
\item Use Double Diffusion Maps and Latent Harmonics to lift the predicted points back to the ambient space: $\{\boldsymbol{\tilde{\phi}}_i\}_{i=1}^{N^\star} \rightarrow \{\tilde{\textbf{x}}_i\}_{i=1}^{N^\star} \sim f_{\textbf{x}}(\cdot)$
\end{itemize}

\end{algorithmic}
\end{algorithm}

\section{Numerical Examples}
\label{sec:examples}
\noindent
The effectiveness of the proposed method is illustrated by two examples. The first  example involves the S-shaped ribbon data structure discussed in Section \ref{sec:problem}. The second example is a composite material system. For each example, samples are generated using the  m-SGM1 (resp. m-SGM2) method (where the number refers to the way we calculate the score function on the manifold: 1 for mini-batch MCS and 2 for Score-matching), as well as using the  m-PLoM approach. In our work,
the Python library datafold \cite{lehmberg2020datafold} was used for the Diffusion Maps and the associated local-linear regression algorithms. We remind the reader of the following notations: raw data are denoted by $\textbf{x}$, latent variables are denoted by  $\boldsymbol{\Phi}$ (with coordinates $\boldsymbol{\phi}_{i}$).

\subsection{Example 1: S-shaped dataset}

\begin{figure}[!htb]
    \centering
    \begin{subfigure}[b]{0.49\textwidth}
        \includegraphics[width=\textwidth]{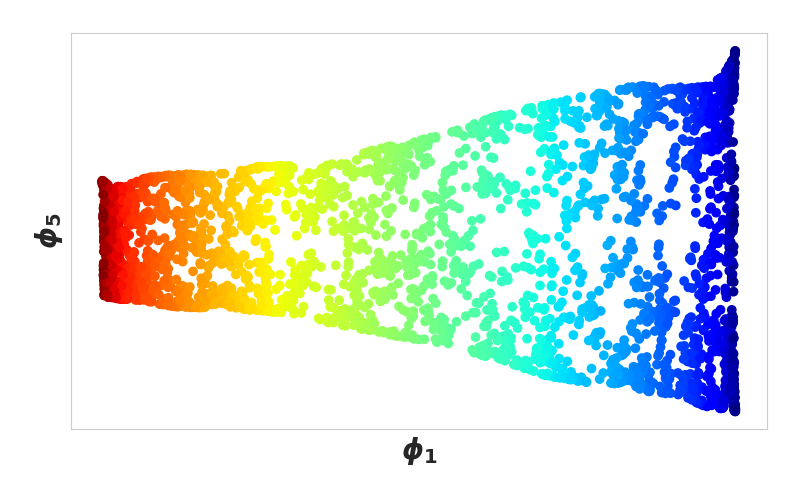}
        \caption{}
        \label{fig:S_curve_dmaps}
    \end{subfigure}
    \hfill 
    \begin{subfigure}[b]{0.49\textwidth}
        \includegraphics[width=\textwidth]{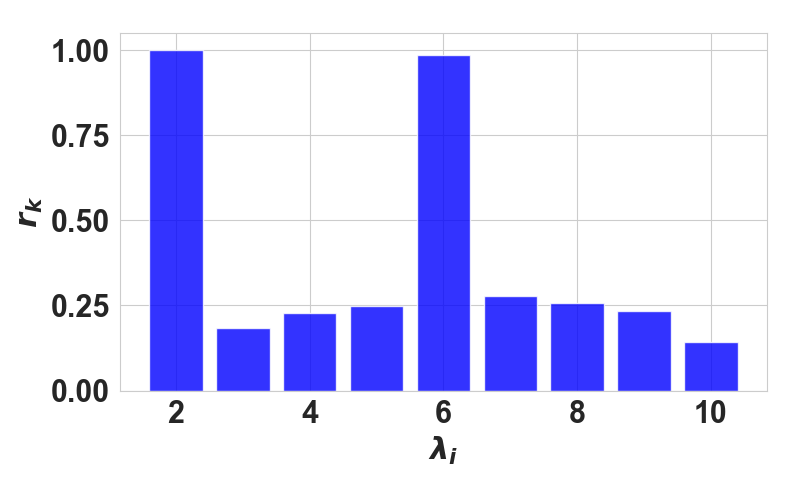}
        \caption{}
        \label{fig:S_curve_residuals}
    \end{subfigure}
    
    \caption{ (a) Diffusion Maps coordinates $\boldsymbol{\phi}_1, \boldsymbol{\phi}_5$ of the three-dimensional dataset. (b)  The residual $r_k$ indicates
  that $\boldsymbol{\phi}_1, \boldsymbol{\phi}_5$ are the two non-harmonic coordinates.}
    \label{fig:S3D_true_data}
\end{figure}

\noindent
In the first example we want to generate additional realizations of the S-shaped ribbon dataset (with $N=3,000$ points) discussed in Section \ref{sec:problem}.  The first step involves applying dimensionality reduction with Diffusion Maps to identify the parsimonious coordinates that parameterize the data. Figure~\ref{fig:S_curve_dmaps} shows the two non-harmonic diffusion coordinates, $\boldsymbol{\Phi}=\{\boldsymbol{\phi}_1, \boldsymbol{\phi}_5\}$, which capture the intrinsic two-dimensional structure of the data set. This is corroborated by the larger residuals $r_k$ of the local
linear regression algorithm proposed by Dsilva et al. \cite{dsilva2018parsimonious}
shown in Fig.~\ref{fig:S_curve_residuals}. The next step involves using Double Diffusion Maps to learn the mapping $\boldsymbol{\Phi}_{\text{new}} \rightarrow \textbf{x}_{\text{new}}$ via the intermediate coordinates $\boldsymbol{\Psi}_{\text{new}}$ (see \ref{sec:ddmaps}), so that we can lift the generated data back to the ambient space. A key aspect of using the trained regressor is that the samples drawn from $f_{\boldsymbol{\Phi}}(\cdot)$ need to live on the manifold. This is necessary because the Geometric Harmonic (GH) function (see \ref{sec:GH}) exhibit numerical instabilities when used to lift points very close to/outside of the manifold boundaries. 

\begin{figure}[!htb]
    \centering
    \begin{subfigure}[b]{0.48\textwidth}
        \includegraphics[width=\textwidth]{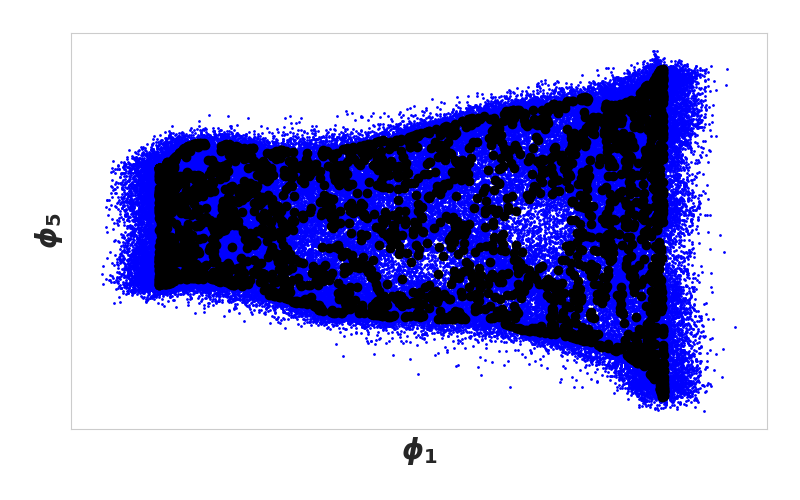}
        \caption{}
        \label{fig:S_mSGM_dmaps_prediction}
    \end{subfigure}
    \hfill
    \begin{subfigure}[b]{0.48\textwidth}
        \includegraphics[width=\textwidth]{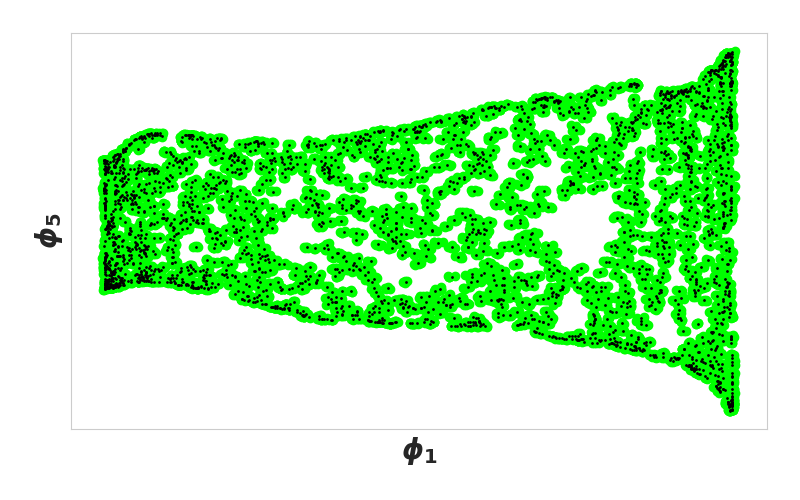}
        \caption{}
        \label{fig:S_mSGM_dmaps_selection}
    \end{subfigure}
    
    \caption{100,000 generated points (blue)  in the latent space using: (a) m-SGM1 (same behavior is observed for m-SGM2). The black points correspond to diffusion maps coordinates of the original dataset. (b) 30,000 selected points (green) using KDTree with 10 nearest neighbors.}
    \label{fig:S3D_true_data}
\end{figure}

In the m-SGM frameworks, as we increase the number of generated  samples, this condition will not be satisfied due to the nature of the SDE, generating samples that  extend beyond the manifold edges. To address this challenge, we propose the following steps: First, we generate a large number $N_s$ of samples from $f_{\boldsymbol{\Phi}}(\cdot)$ and then, for each point in the original dataset, we select the $n$ nearest neighbors among the generated points. This way we ensure that the selected generated points will live on (or very close to) the low-dimensional manifold.  To find the nearest neighbors, we use a $k$-dimensional tree (KDTree), which is a binary search tree used to organize points \cite{ram2019revisiting}. Figure \ref{fig:S_mSGM_dmaps_prediction} depicts 100,000  initial generated points in the latent space using the m-SGM1 approach (similar results are obtained with the m-SGM2 method). We can see that the a portion of the generated points live outside the boundaries of the manifold. Figure~\ref{fig:S_mSGM_dmaps_selection} shows those generated points that were selected using KDTree with 10 nearest neighbors. This results in $N_t = n \times N = 10 \times 3,000 = 30,000$  points sampled from $f_{\boldsymbol{\Phi}}(\cdot)$. 


\begin{figure}[!htb]
    \centering
    \begin{subfigure}[b]{0.49\textwidth}
        \includegraphics[width=\textwidth]{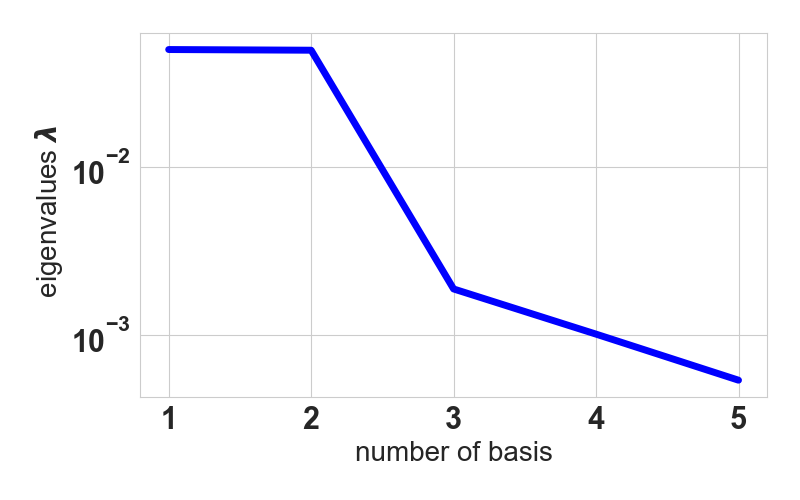}
        \caption{}
        \label{fig:S_curve_mplom_eigs}
    \end{subfigure}
    \hfill 
    \begin{subfigure}[b]{0.49\textwidth}
        \includegraphics[width=\textwidth]{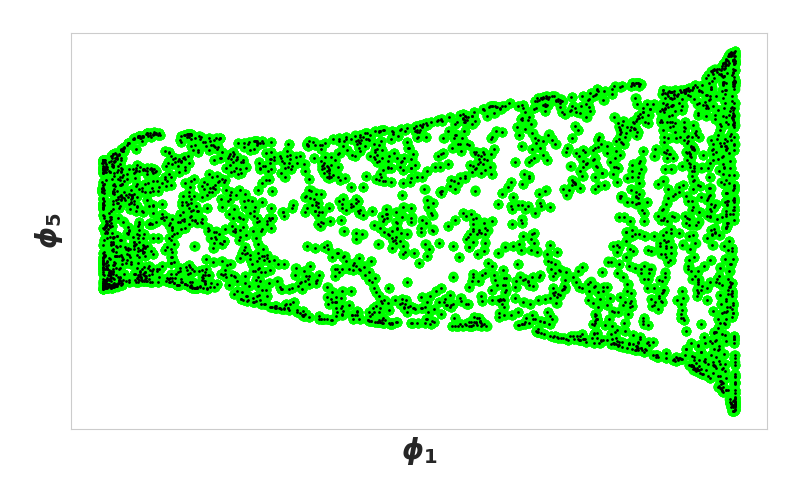}
        \caption{}
        \label{fig:S_curve_mplom_dmaps_prediction}
    \end{subfigure}
    
    \caption{ (a) Eigenvalues of the diffusion matrix  of the PCA in PLoM. (b)  30,000 generated points (green) using PLoM.}
    \label{fig:S3D_mplom_data}
\end{figure}

Next, we sample from $f_{\boldsymbol{\Phi}}(\cdot)$ using the m-PLoM approach. To solve the It\^o equations we  use a Störmer-Verlet algorithm; we use a value of $f_0$ equal to 1, and an
integration step $\Delta_r$ of 0.0005 (see \ref{sec:plom}). Figure \ref{fig:S_curve_mplom_eigs} shows the decay of the eigenvalues of the diffusion matrix associated with the (optimal value of the) bandwidth
parameter $\epsilon=10$. It is clear from this figure that the eigenvalues are grouped in two clusters; the leading cluster contains two eigenvalues, suggesting a two-dimensional parameterization of the data. Figure \ref{fig:S_curve_mplom_dmaps_prediction} depicts 30,000 additional realizations (green) of the latent coordinates generated by the first PLoM step.

\begin{figure}[!htb]
    \centering
    \begin{subfigure}[b]{0.32\textwidth}
        \includegraphics[width=\textwidth, trim={1cm 3cm 1cm 2cm}, clip]{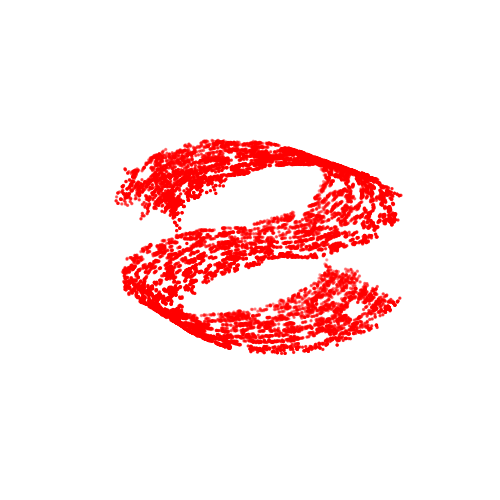}
        \caption{m-SGM1}
        \label{fig:s_data_SGM}
    \end{subfigure}
    \hfill 
    \begin{subfigure}[b]{0.32\textwidth}
        \includegraphics[width=\textwidth, trim={1cm 3cm 1cm 2cm}, clip]{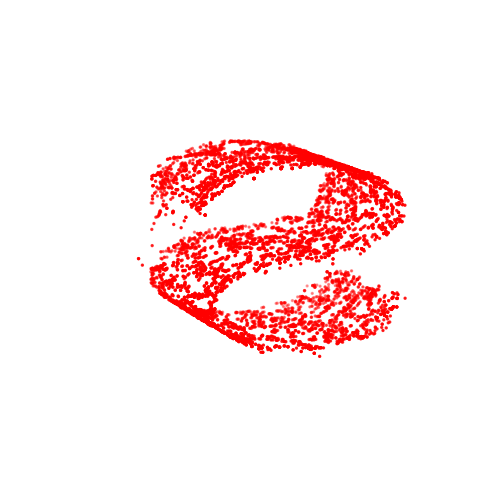}
        \caption{m-SGM2}
        \label{fig:s_data_mSGM}
    \end{subfigure}
    \hfill 
    \begin{subfigure}[b]{0.32\textwidth}
        \includegraphics[width=\textwidth, trim={1cm 3cm 1cm 2cm}, clip]{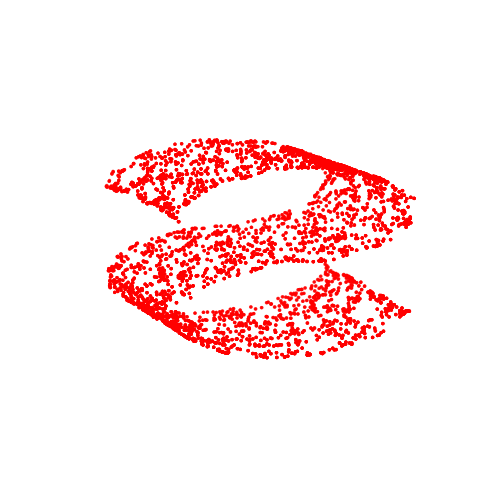}
        \caption{m-PLoM}
        \label{fig:s_data_mplom}
    \end{subfigure}

    \caption{Comparison of the three approaches for generating points on the S-shaped ribbon.}
    \label{fig:generative_learning_comparison}
\end{figure}

\begin{figure}[!htb]
    \centering
    \begin{subfigure}[b]{0.8\textwidth}
        \includegraphics[width=\textwidth]{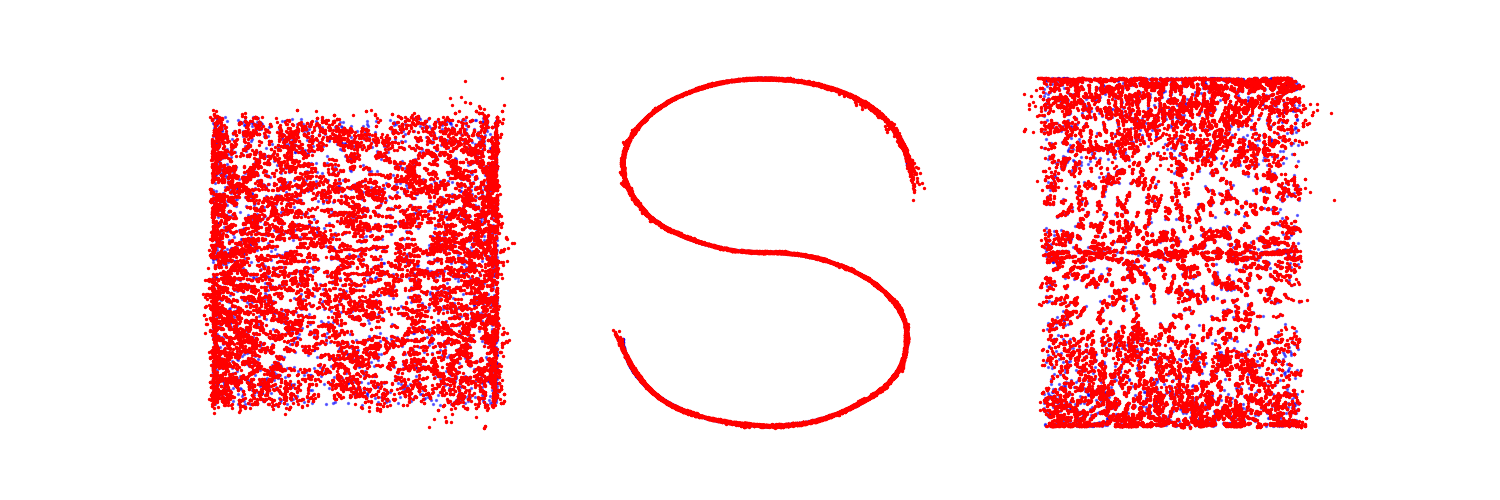}
        \caption{m-SGM1}
        \label{fig:s2d_data_1mSGM}
    \end{subfigure}
    
        \begin{subfigure}[b]{0.8\textwidth}
        \includegraphics[width=\textwidth]{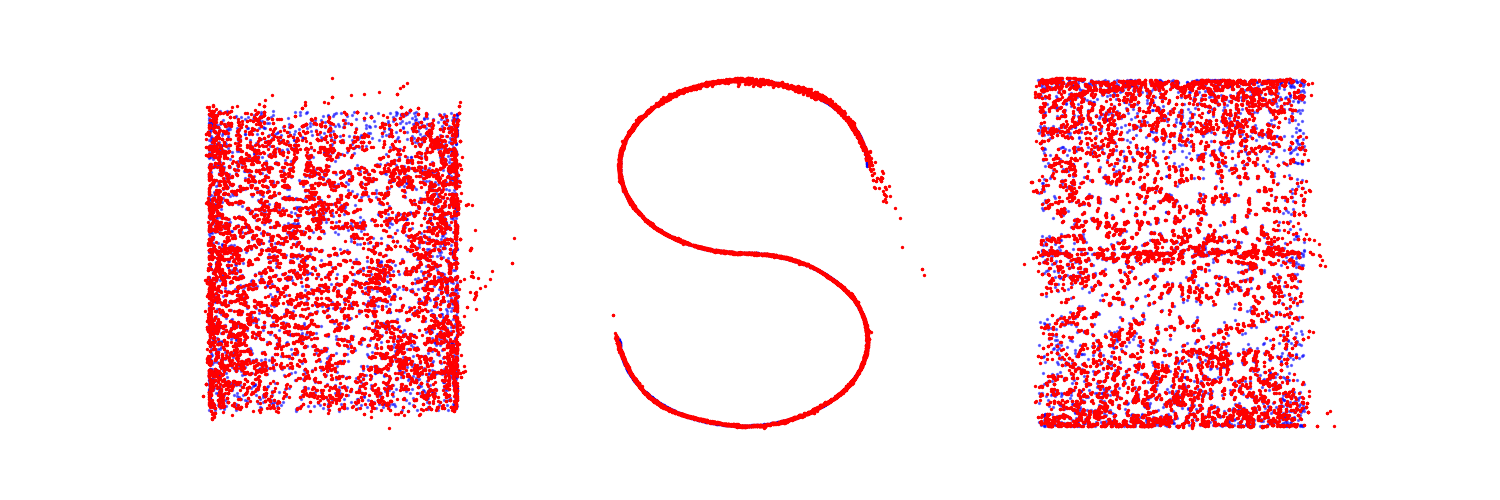}
        \caption{m-SGM2}
        \label{fig:s2d_data_2mSGM}
    \end{subfigure}
    
    \begin{subfigure}[b]{0.8\textwidth}
        \includegraphics[width=\textwidth]{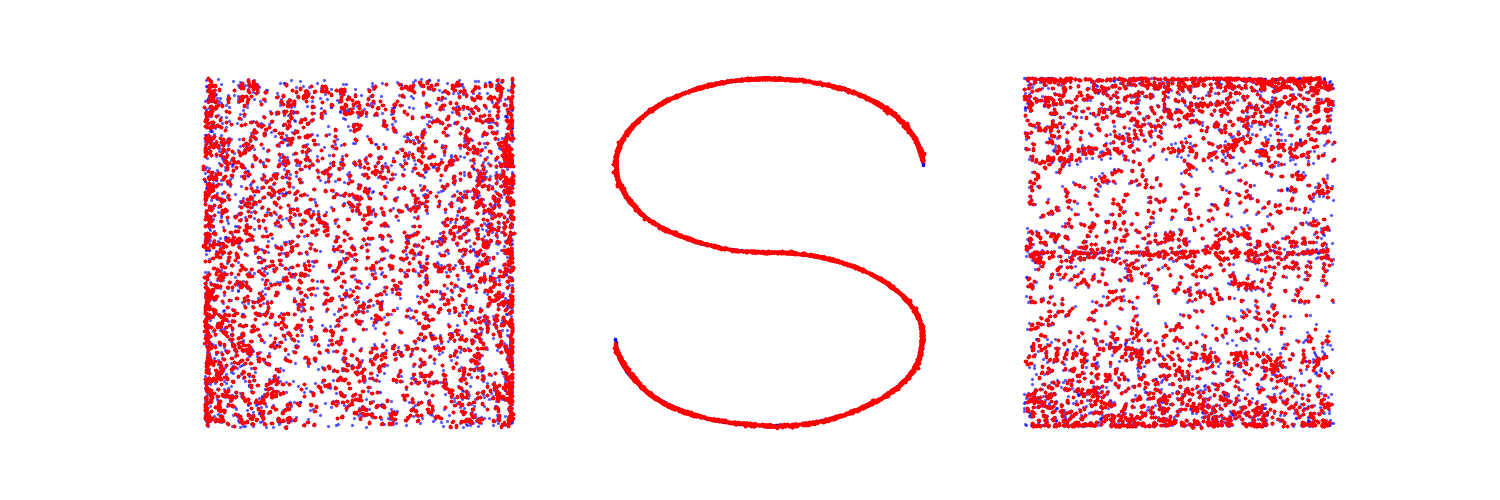}
        \caption{m-PLoM}
        \label{fig:s2d_data_mplom}
    \end{subfigure}

    \caption{Further comparison of the three methods projected along different axis pairs: (a)  m-SGM1, (b) m-SGM2, and (c) m-PLoM.}
    \label{fig:generative_2d_projections}
\end{figure}

\begin{figure}[!htb]
    \centering
    \begin{subfigure}[b]{0.85\textwidth}
        \includegraphics[width=\textwidth]{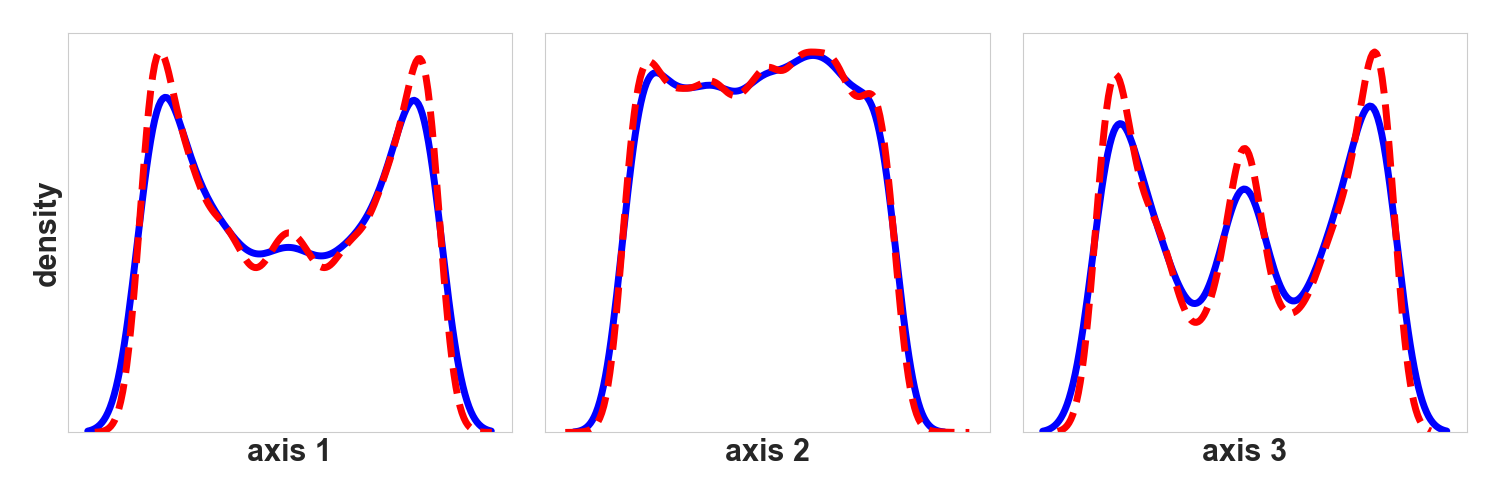}
        \caption{m-SGM1}
        \label{fig:s_pdf_data_1mSGM}
    \end{subfigure}
    
    \begin{subfigure}[b]{0.85\textwidth}
        \includegraphics[width=\textwidth]{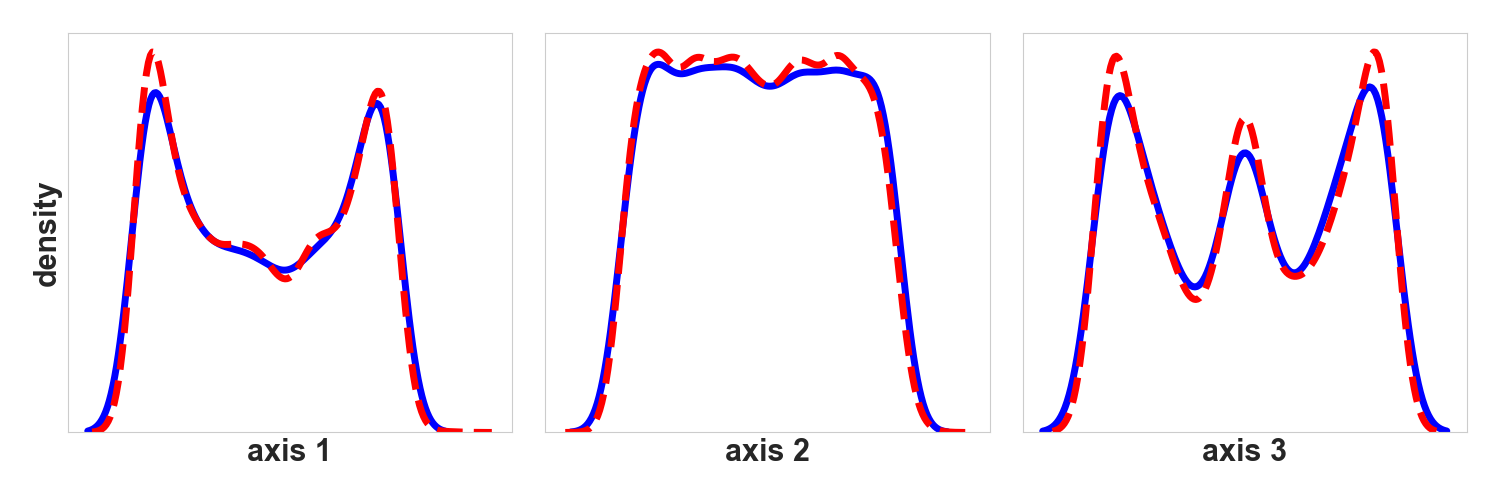}
        \caption{m-SGM2}
        \label{fig:s_pdf_data_2mSGM}
    \end{subfigure}
    
    \begin{subfigure}[b]{0.85\textwidth}
        \includegraphics[width=\textwidth]{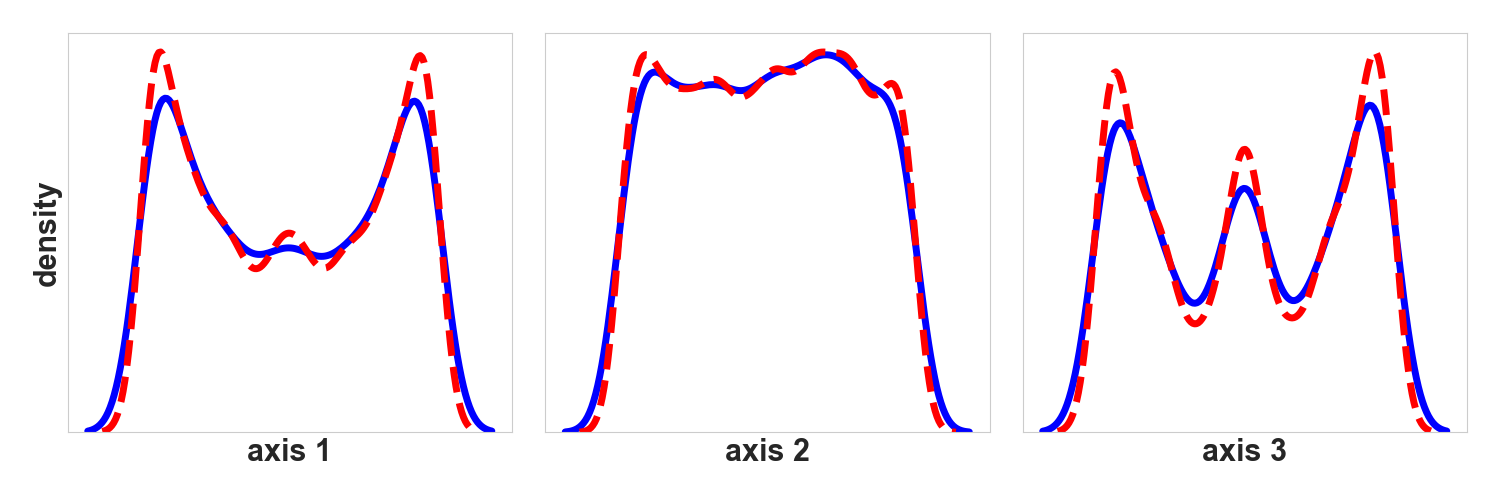}
        \caption{m-PLoM}
        \label{fig:s_pdf_data_mplom}
    \end{subfigure}

    \caption{Comparison of the generated marginal densities (red dashed lines) against the ground truth densities (blue solid lines) across the three axis.}
    \label{fig:s_DENSITIES}
\end{figure}

Figure~\ref{fig:generative_learning_comparison} provides a three-dimensional visualization of the generated samples using the m-SGM and m-PLoM approaches. In Fig.~\ref{fig:s_data_SGM}, the data generated with m-SGM1 are depicted, revealing that the generated points respect the underlying manifold structure. Figure~\ref{fig:s_data_mSGM} demonstrates the performance of m-SGM2, which also aligns the generated points with the ground truth, accurately capturing the underlying manifold structure. Finally, Fig.
~\ref{fig:s_data_mplom} shows m-PLoM which also accurately generates points that respect the manifold structure of the dataset.  The effectiveness of each method is further illustrated in Figure~\ref{fig:generative_2d_projections}, where we can see the projections of the generated points along the three axis pairs. 
Figures \ref{fig:s2d_data_1mSGM} and \ref{fig:s2d_data_2mSGM} show the m-SGM, where the generated data aligns with the underlying structure. 
Figure \ref{fig:s2d_data_mplom} supports the finding that m-PLoM also achieves a good alignment with the true manifold. In all figures, the red points represent the generated data, while the blue points represent the original dataset. Last but not least, Fig.~\ref{fig:s_DENSITIES} provides a comparison of the marginal densities for m-SGM and m-PLoM approaches. The blue lines represent the true density, while the red dashed lines depict the  density of the generated data. 
More specifically, we see that that the densities estimated with in all cases are well aligned with the true distributions.

\subsection{Example 2: Multiscale Modeling of a Material System}

\noindent
In the second example, we consider an eight-laminae composite laminate whose performance is described in terms of its full stress-strain curve as measured at the scale of the entire laminate \cite{ghanem2022probabilistic}. The laminate comprises eight laminae constructed from continuous noncrimp fabric (NCF), with resin occupying the spaces between the fibers. Each lamina contains continuous tows oriented in a single direction. The laminae are arranged in a 
\([0^\circ/45^\circ/-45^\circ/90^\circ/90^\circ/-45^\circ/45^\circ/0^\circ]\) 
configuration to achieve quasi-isotropic symmetry in the laminate structure. The carbon fiber tows consist of 12,000 fibers (T700SC 12000 50C). Due to the challenges of directly measuring the mechanical and geometric properties of microconstituents \textit{in situ}, we utilize a computational model to generate observables that are governed by underlying physical principles. This physics spans four hierarchical length scales: laminate, lamina, tow, and fiber.

\begin{figure}[!htb]
    \centering
    \begin{subfigure}[b]{0.49\textwidth}
        \includegraphics[width=\textwidth]{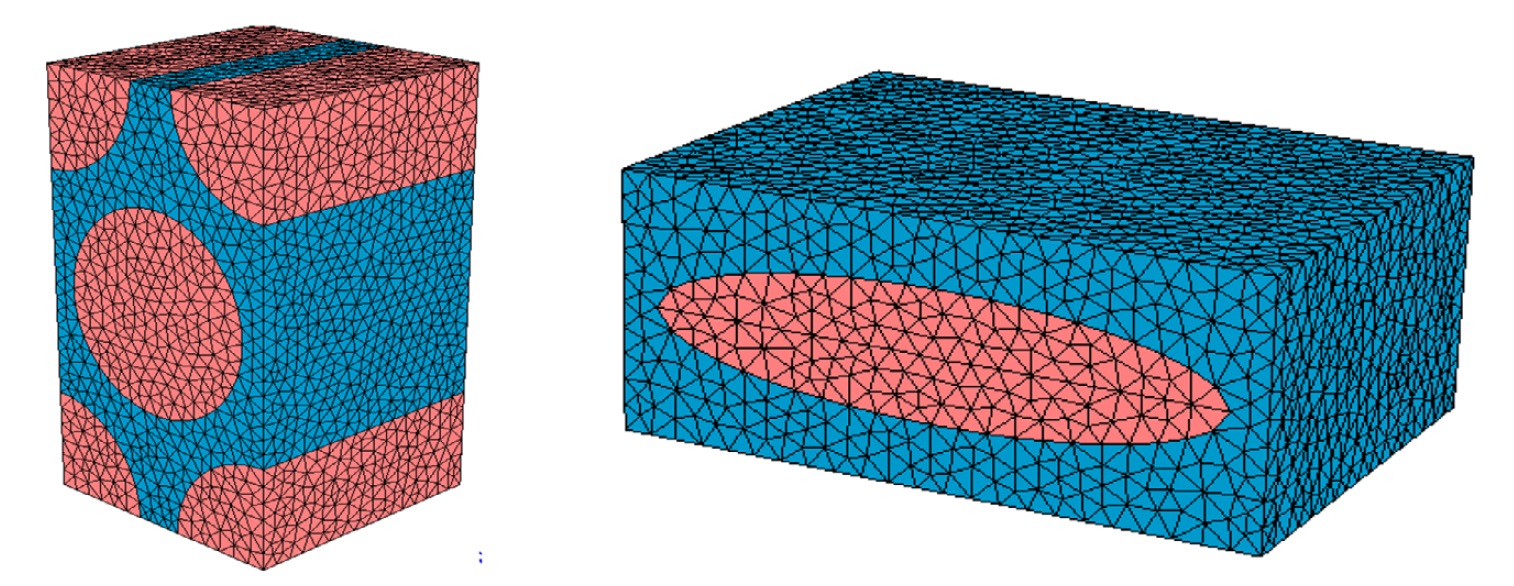}
        \caption{}
        \label{fig:material_sample}
    \end{subfigure}
    \hfill 
    \begin{subfigure}[b]{0.49\textwidth}
        \includegraphics[width=\textwidth]{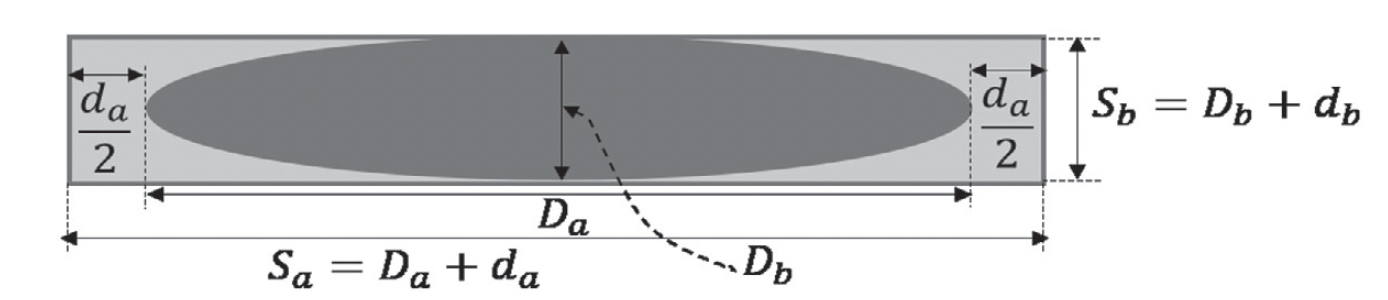}
        \caption{}
        \label{fig:material_NCF}
    \end{subfigure}
    
    \caption{ (a) A volumetric unit cell within
a tow (left) and within a NCF lamina (right).
(b) A schematic drawing of a unit cell in a NCF unidirectional lamina. (NCF: noncrimp fabric).}
    \label{fig:meterial_data}
\end{figure}

A lamina is composed of ellipsoidal tows embedded within a resin matrix, with each tow consisting of a bundle of circular fibers, which are themselves infused into the resin matrix. Significant nonlinear interactions can arise between features, behaviors, and scales. To account for these nonlinearities, an upscaling methodology is employed using the software Multiscale Designer \cite{wollschlager2021technical}. Figure \ref{fig:material_sample} depicts the volumetric unit cell which comprises a unidirectional tow with an elliptic cross-section surrounded by resin.

The inputs to the finest-scale simulation include the mechanical properties of the fibers (denoted with subscript $f$) and resin (denoted with subscript $m$), \(\textbf{P}^0_f\) and \(\textbf{P}^0_m\), respectively. These input parameters for the microscale simulator are collectively represented as:
\[
\textbf{P}^0 = (V_f^{F, T}, E_{f, A}, E_{f, T}, G_{f, A}, \nu_{f, A}, \nu_{f, T}, E_m, \nu_m),
\]
where \(V_f^{F, T}\) denotes the fiber volume fraction, \(T_f\) represents the fiber tensile strength, \(E_f\) and \(G_f\) are the fiber elastic modulus and shear modulus, \(\nu_f\) is the fiber Poisson's ratio (with \(A\) and \(T\) subscripts denoting axial and transverse directions, respectively), and \(E_m\) and \(\nu_m\) are the matrix elastic modulus and Poisson's ratio.

The second-scale simulation, which evaluates the properties of the lamina, uses as input the effective mechanical properties of the tows along with geometric parameters that define the major and minor axes of the tows and the edge-to-edge spacing between these ellipsoidal structures. The geometric parameters define the dimensions of the unit cell of an NCF unidirectional lamina, as illustrated in Fig.~\ref{fig:material_NCF} where \(D_a\) and \(D_b\) represent the diameters of the tow along the major and minor axes, respectively, while \(d_a\) and \(d_b\) denote the gaps between the tows in these directions, filled by the matrix. Observations from micrographs show that \(d_b\) is very small; thus, for finite element discretization, this parameter is assumed constant at \(0.06\,\text{mm}\). The unit cell dimensions are \(S_a = D_a + d_a\) and \(S_b = D_b + d_b\). Micrographs also indicate that the tow cross-sections can be approximated as ellipsoidal shapes, though detailed validation is beyond this work's scope. The input parameters \(\textbf{P}^1\) consist of the geometry parameters \(\textbf{P}^1_g\) describing the unit cell in an NCF lamina, and the material properties \(\textbf{P}^1_m\) of the isotropic matrix. Thus, \(\textbf{P}^1 = (D_a, d_a, D_b, E_m, \nu_m)\). The upscaling process then computes the homogenized constitutive properties of an NCF unit cell in a lamina. The Multiscale Designer software employs hierarchical reduced-order homogenization. The reader is referred to \cite{ghanem2022probabilistic} for the details of the analysis.

\begin{table}[ht]
    \centering
    \caption{Input parameters for numerical simulations of microscale simulations}
    \begin{tabularx}{\linewidth}{@{}llll@{}}
        \toprule
        \textbf{Property} & \textbf{Variable} & \textbf{Definition} & \textbf{Range \& COV} \\ 
        \midrule
         & $E_{f_a}$ & Fiber axial modulus (GPa) & [172--206], 4.5 \\
         & $E_{f_t}$ & Fiber transverse modulus (GPa) & [12.5--16.5], 6.9 \\
        & $G_{f_a}$ & Fiber shear modulus (GPa) & [7.3--9.7], 6.4 \\
         & $\nu_{f_a}$ & Fiber axial Poisson ratio & [0.29--0.38], 6.72 \\
         \textbf{Fiber}  & $\nu_{f_t}$ & Fiber transverse Poisson ratio & [0.17--0.23], 7.5 \\
        & $\rho_{f_ac}$ & Compression modulus/axial modulus & [0.72--0.88], 0.05 \\
        & $\sigma_{yf_a}$ & Fiber axial yield strength (GPa) & [2.6--4.0], 10.6 \\
         & $\rho_{f_t} = \sigma_{yt}/\sigma_{ya}$ & Transverse strength/axial strength & [0.02--0.06], 0.025 \\
        & $\rho_{f_c} = \sigma_{yac}/\sigma_{ya}$ & Compressive strength/axial strength & [0.55--0.65], 0.0417 \\
        \midrule
        & $\nu_m$ & Resin Poisson ratio & [0.32--0.4], 0.0556 \\
       &  $\sigma_{ym}$ & Mean stress at damage initiation (MPa) & [27--29], 1.97 \\
        & $K_{0m}$ & Yield strength (MPa) & [31.6--33.6], 1.53 \\
         \textbf{Resin}& $K_{1m}$ & Ultimate strength (MPa) & [52--54], 0.94 \\
         & $H_d$ & Linear term for hardening law & [0.0033--0.0035], 0.0094 \\
         & $\delta$ & Exponent for evolution law & [40--44], 1.47 \\
        \midrule
         & $D_a$ & Major axis of tow's ellipsoid & Figure \ref{fig:meterial_data2}  \\
        \textbf{Tow} & $D_b$ & Minor axis of tow's ellipsoid & --  \\
         & $d_a$ & Spacing between tows & Empirical distribution  \\
        \bottomrule
    \end{tabularx}
    \label{tab:1a}
\end{table}

To incorporate uncertainties in the configuration and material properties, the variables in Table \ref{tab:1a} are modeled as random variables. In this work, 7 out of the 18 random variables were treated as observables: \(E_{f_t}\), \(\rho_{f_ac}\), \(\sigma_{yf_a}\), \(\rho_{f_t}\), \(D_a\), \(D_b\), and \(d_a\). The remaining variables were not assumed to be deterministic, but were left unspecified, thereby increasing the uncertainty in probabilistic inferences and broadening the associated pdf.

\begin{figure}[!htb]
    \centering
    \begin{subfigure}[b]{0.49\textwidth}
        \includegraphics[width=\textwidth]{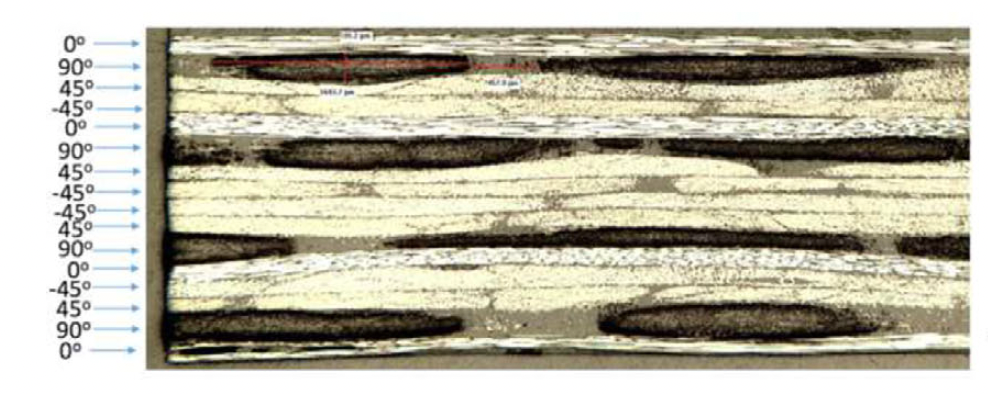}
        \caption{}
        \label{fig:material_sample1}
    \end{subfigure}
    \hfill 
    \begin{subfigure}[b]{0.49\textwidth}
        \includegraphics[width=\textwidth]{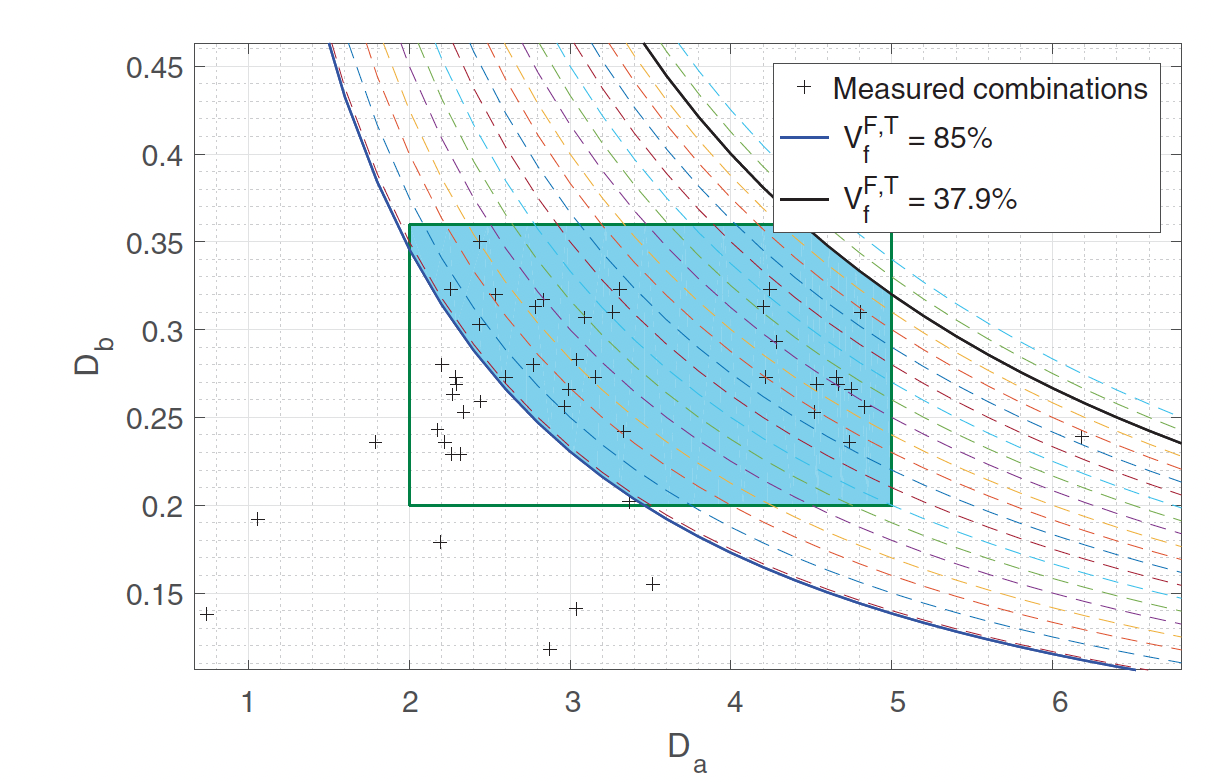}
        \caption{}
        \label{fig:material_Da}
    \end{subfigure}
    
    \caption{ (a) Micrograph of tow and laminae.
(b) Support of joint probability density function of tow
diameters.}
    \label{fig:meterial_data2}
\end{figure}

Except for geometry parameters, all variables are assumed to be statistically independent, following a symmetric beta distribution (shape parameter 1.5) with ranges defined by manufacturer specifications and expert opinions. The parameters of the tow geometry (\(D_a\), \(D_b\), \(d_a\)) are modeled using experimental data. A joint pdf is defined for \(D_a\) and \(D_b\), constrained by the lamina-scale volume fraction, while \(d_a\) is independently modeled. The joint pdf support, shown as a blue-shaded area in Fig.\ref{fig:material_Da}, incorporates experimental data points, with deviations attributed to the irregular geometry of real tows, as seen in the micrograph in Fig.~\ref{fig:material_sample1}.

The model outputs are represented as stress-strain curves through the inelastic regime, generated under two loading conditions: tension and three-point bending tests. In addition, 14 tensile and 11 bending experiments were conducted in the laboratory, with stress-strain histories recorded. No further microstructural data for the constituents were available for these tests.

Our dataset consists of 800 points (simulations) in $\mathbb{R}^{449}$, where the 449 dimensions are structured as follows: dimensions 1–7 represent microscale properties, dimensions 8–143 correspond to stress values from tensile tests at predefined strain levels, and dimensions 144–449 correspond to stress values from bending tests at predefined strain levels. In this scenario, vanilla SGM will not provide reliable results since 800 samples are not sufficient for the MCS-based estimation of the score function  (curse of dimensionality). Consequently, the results will have significant statistical uncertainty. The performance of PLoM was well-studied in \cite{ghanem2022probabilistic}. In this work we utilize m-SGM1 and m-PLoM to generate additional realizations of the 49-dimensional vectors. We expect that m-SGM2 will give similar results.

\begin{figure}[!htb]
    \centering
    \begin{subfigure}[b]{0.49\textwidth}
        \includegraphics[width=\textwidth]{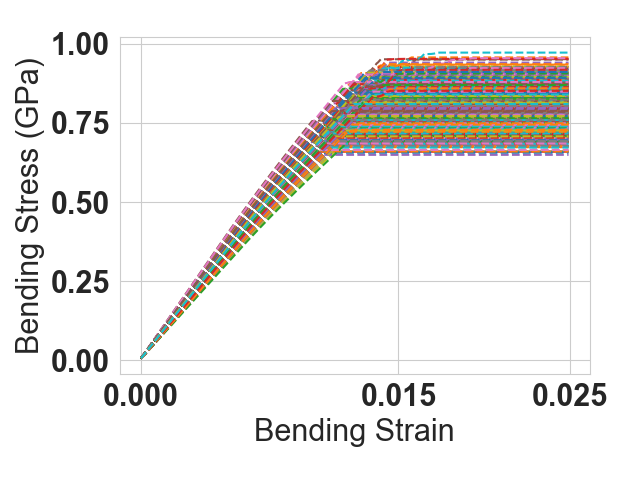}
        \caption{}
        \label{fig:material_bending_train}
    \end{subfigure}
    \hfill 
    \begin{subfigure}[b]{0.49\textwidth}
        \includegraphics[width=\textwidth]{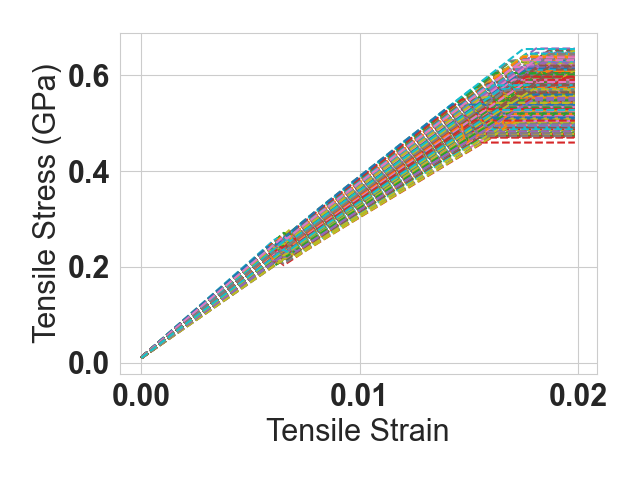}
        \caption{}
        \label{fig:material_tensile_train}
    \end{subfigure}
    
    \caption{  Training datasets for: (a) bending model and (b) tension model.}
    \label{fig:strain-stress-train}
\end{figure}

Figure \ref{fig:strain-stress-train} illustrates the stress-strain components from the training dataset, displaying curves for both bending (Fig.~\ref{fig:material_bending_train}) and tensile (Fig.~\ref{fig:material_tensile_train}) loading configurations. These curves exhibit a similar range and overall behavior. The complete set of observables includes not only the tensile and flexural curves but also microscale constituents (mechanical properties of fibers and resin) and mesoscale constituents (tow diameters and separations).

\begin{figure}[!htb]
    \centering
    \begin{subfigure}[b]{0.49\textwidth}
        \includegraphics[width=\textwidth]{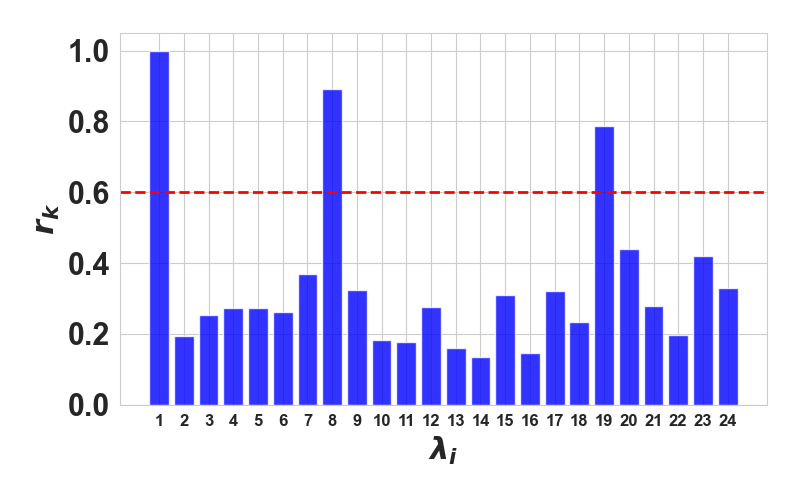}
        \caption{}
        \label{fig:dmaps_msgm-eigs}
    \end{subfigure}
    \hfill 
    \begin{subfigure}[b]{0.49\textwidth}
        \includegraphics[width=\textwidth]{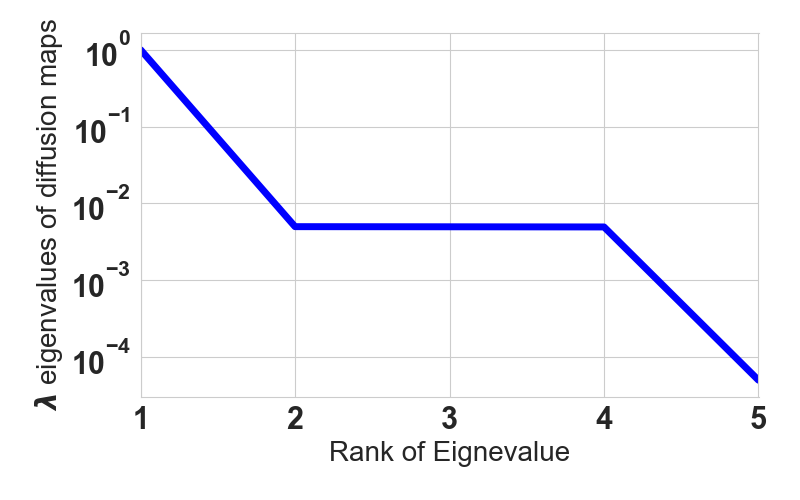}
        \caption{}
        \label{fig:dmaps_mplom_eigs}
    \end{subfigure}
    
    \caption{(a) Residual $r_k$ of the local linear regression algorithm for the diffusion maps, indicating that coordinates $\phi_1, \phi_8$ and $\phi_{19}$ are non-harmonics. (b) Decay of the eigenvalues of the diffusion matrix in m-PLoM.}
    \label{fig:material_ddmaps_eigs}
\end{figure}

Figure \ref{fig:dmaps_msgm-eigs} shows the residual $r_k$ of the local linear regression in the difusion maps coordinates for $\epsilon=2$. It is clear from this figure that the three non-harmonic eigenvectors are $\phi_1, \phi_8$ and $\phi_{19}$. Figure \ref{fig:dmaps_mplom_eigs} depicts  the decay of the eigenvalues of the diffusion matrix in PLoM. 

\begin{figure}[!htb]
    \centering
    \begin{subfigure}[b]{0.24\textwidth}
        \includegraphics[width=\textwidth]{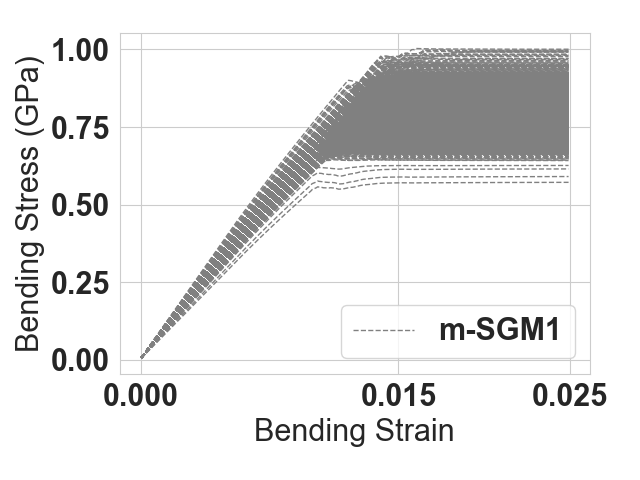}
        \caption{}
        \label{fig:material_bending_mSGM}
    \end{subfigure}
    \hfill 
    \begin{subfigure}[b]{0.24\textwidth}
        \includegraphics[width=\textwidth]{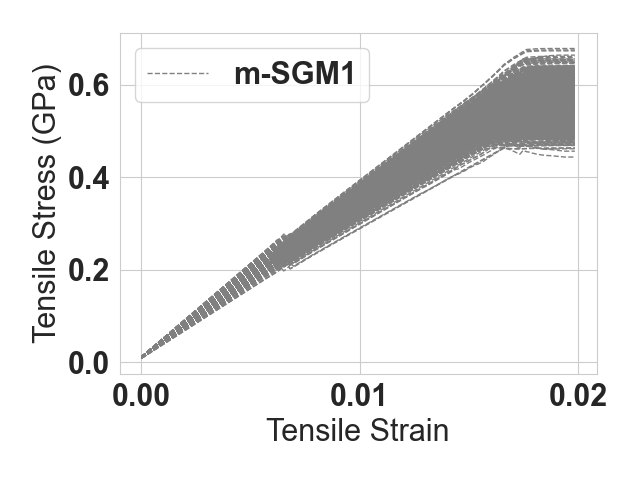}
        \caption{}
        \label{fig:material_tensile_mSGM}
    \end{subfigure}
    \hfill
        \begin{subfigure}[b]{0.24\textwidth}
        \includegraphics[width=\textwidth]{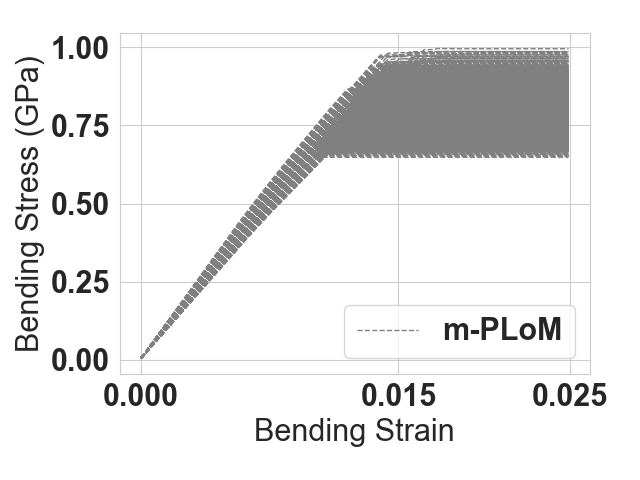}
        \caption{}
        \label{fig:material_bending_mPLoM}
    \end{subfigure}
    \begin{subfigure}[b]{0.24\textwidth}
        \includegraphics[width=\textwidth]{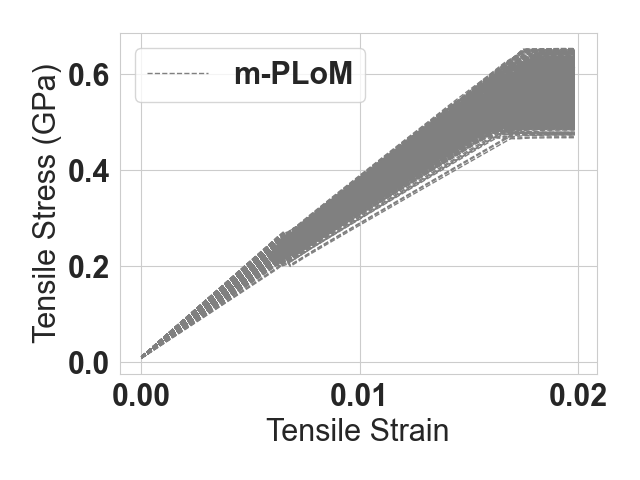}
        \caption{}
        \label{fig:material_tensile_mPLoM}
    \end{subfigure}
    
    \caption{Generated stress-strain curves using: (1) m-SGM1 for: (a) the bending model and (b) the tension model. (2) m-PLoM for: (c) the bending model and (d) the tension model.}
    \label{fig:strain-stress-mSGM}
\end{figure}

Figures \ref{fig:material_bending_mSGM} and  \ref{fig:material_tensile_mSGM} show the 4,000 stress-strain components for the tensile and fluxural loading configurations, respectively, generated using m-SGM1 method. Similarly, Figs.~\ref{fig:material_bending_mPLoM} and \ref{fig:material_tensile_mPLoM} depict the same results obtained using m-PLoM. From these figures we can see that the generated curves are similar to the training set for both methods and for both loading scenarios.

\begin{figure}[!htb]
    \centering
    \begin{subfigure}[b]{0.45\textwidth}
        \includegraphics[width=\textwidth]{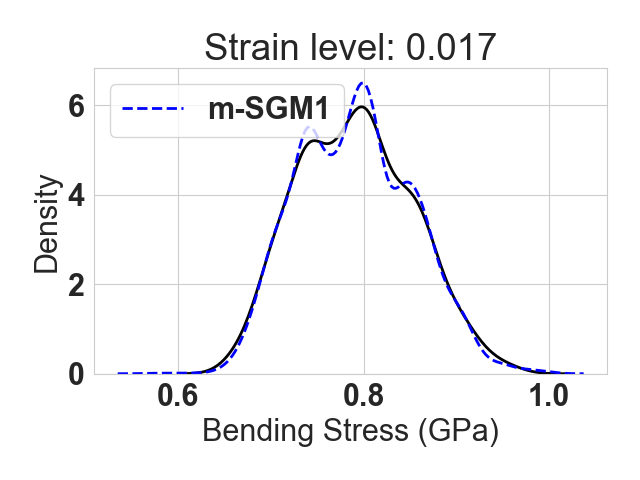}
        \caption{}
        \label{fig:material_bending_0017_mSGM}
    \end{subfigure}
    \hfill 
    \begin{subfigure}[b]{0.45\textwidth}
        \includegraphics[width=\textwidth]{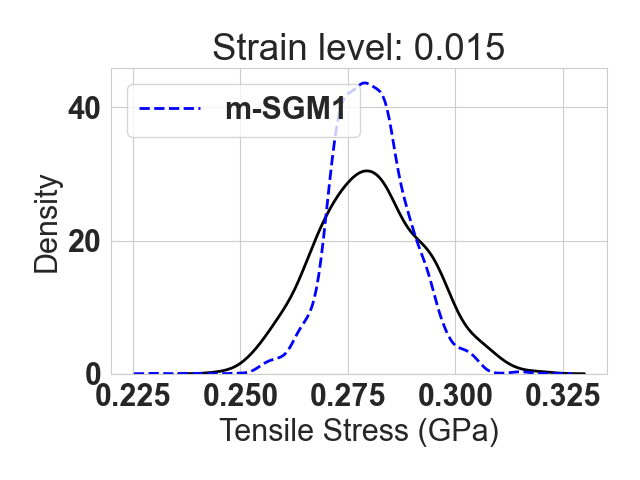}
        \caption{}
        \label{fig:material_tensile_0015_mSGM}
    \end{subfigure}

        \begin{subfigure}[b]{0.45\textwidth}
        \includegraphics[width=\textwidth]{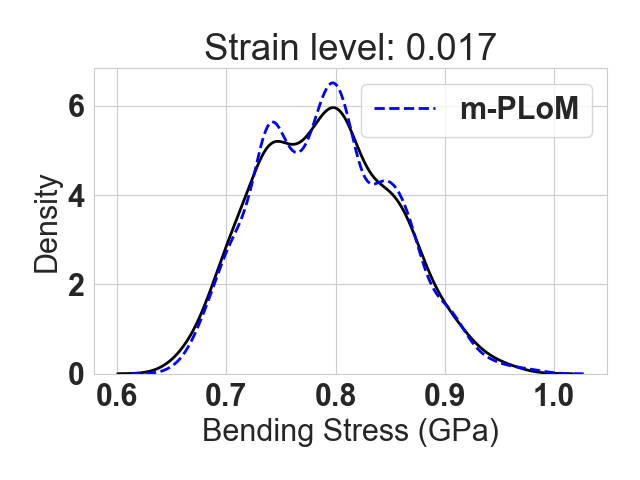}
        \caption{}
        \label{fig:material_bending_0017_mPLoM}
    \end{subfigure}
    \hfill 
    \begin{subfigure}[b]{0.45\textwidth}
        \includegraphics[width=\textwidth]{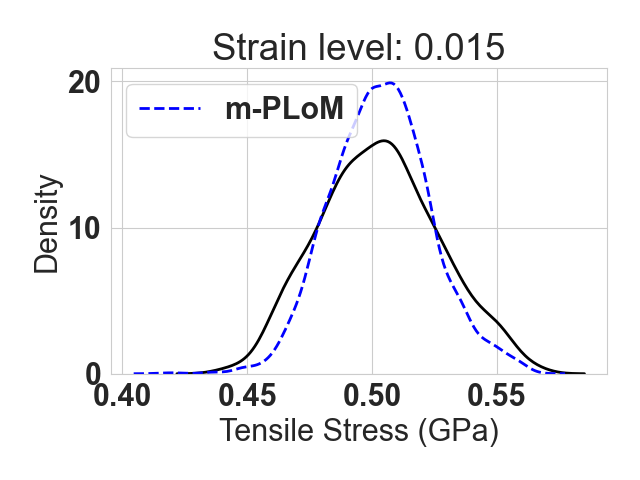}
        \caption{}
        \label{fig:material_tensile_0015_mPLoM}
    \end{subfigure}

    \caption{ Density for bending stress at strain level of 0.0117 for (a) m-SGM1 and (c) m-PLoM; tensile stress for tensile strain level of 0.015 for (b) m-SGM1 and (d) m-PLoM stresses.  }
    \label{fig:pdf-mSGM}
\end{figure}

Figure \ref{fig:pdf-mSGM} shows the density, obtained from the generated data, of the bending stress at a strain level of 0.0117 (Fig.~\ref{fig:material_bending_0017_mSGM}), and the tensile stress at a strain level of 0.015 (Fig.~\ref{fig:material_tensile_0015_mSGM}) for m-SGM1. Similarly,  Fig.~\ref{fig:material_bending_0017_mPLoM} and Fig.~\ref{fig:material_tensile_0015_mPLoM} show the same densities obtained with m-PLoM. The dotted lines in these figures show the density from the generated set while the solid lines show the density from the original dataset.

\begin{figure}[!htb]
    \centering
    \begin{subfigure}[b]{0.49\textwidth}
        \includegraphics[width=\textwidth]{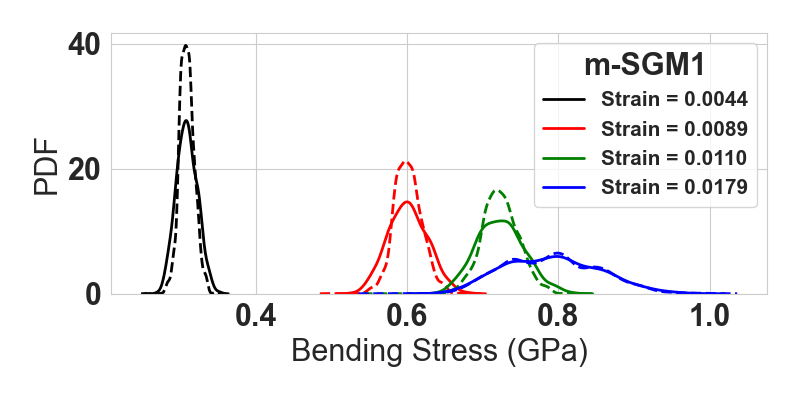}
        \caption{}
        \label{fig:material_bending_all_mSGM}
    \end{subfigure}
    \hfill 
    \begin{subfigure}[b]{0.49\textwidth}
        \includegraphics[width=\textwidth]{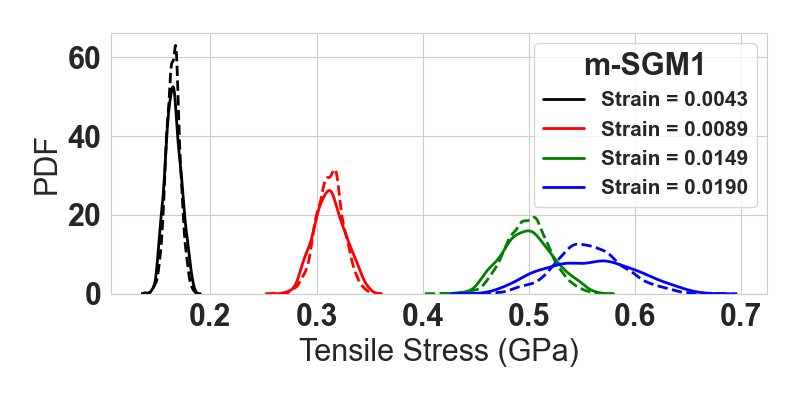}
        \caption{}
        \label{fig:material_tensile_all_mSGM}
    \end{subfigure}

    \vfill

        \begin{subfigure}[b]{0.49\textwidth}
        \includegraphics[width=\textwidth]{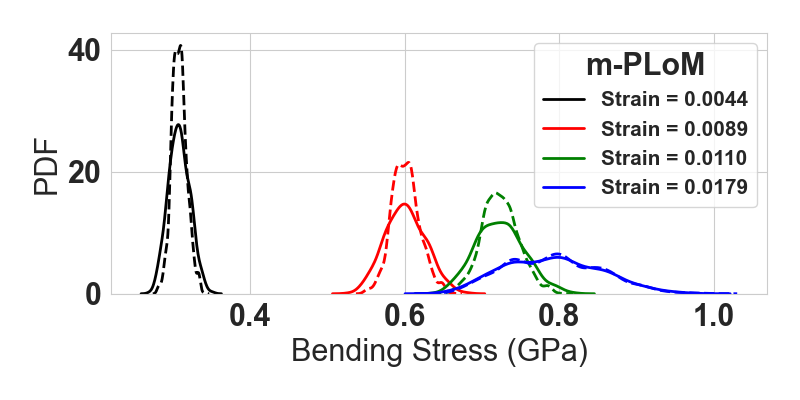}
        \caption{}
        \label{fig:material_bending_all_mPLoM}
    \end{subfigure}
    \hfill 
    \begin{subfigure}[b]{0.49\textwidth}
        \includegraphics[width=\textwidth]{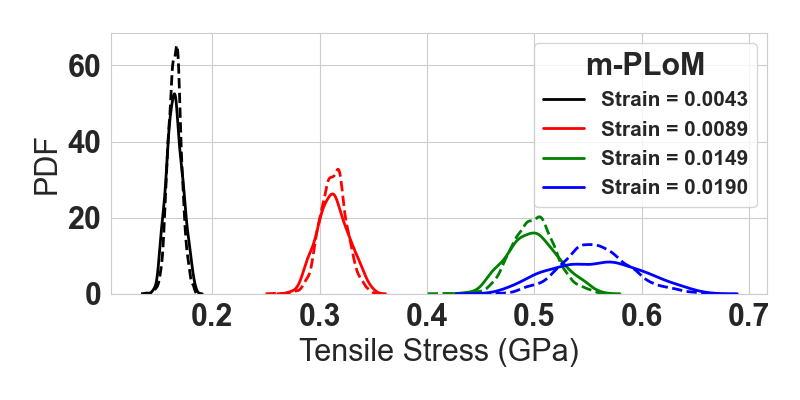}
        \caption{}
        \label{fig:material_tensile_all_mPLoM}
    \end{subfigure}
    
    \caption{  Density of the flexural (a) and (c) and tensile stresses (b) and (d) at different strain levels for m-SGM1 and m-PLoM; leftmost curves are in the linear elastic regime, while the rightmost curves are in the inelastic regime.}
    \label{fig:pdf-all-mSGM}
\end{figure}

Figures \ref{fig:material_bending_all_mSGM},  \ref{fig:material_tensile_all_mSGM}, \ref{fig:material_bending_all_mPLoM}, and  \ref{fig:material_tensile_all_mPLoM} show the density of the generated samples for the bending and tensile stresses, respectively, at different strain levels for m-SGM1 and m-PLoM. Again,
the dotted lines in these figures show the density from the generated set while the solid lines show the density from the original dataset.

\section{Conclusions}
\label{sec:conclusions}
\noindent
In this work, we introduced a generative modeling framework that integrates diffusion models with manifold learning to efficiently sample high-dimensional data densities. Utilizing Diffusion Maps, we identified and utilized a latent low-dimensional structure underlying the data, enabling more efficient sampling. Two distinct approaches for sampling from the latent density were proposed: a score-based diffusion model trained via a neural network and the probabilistic ``learning on the manifold`` solver. The generated samples were subsequently lifted back to the ambient space using Double Diffusion Maps, facilitating the reconstruction of high-dimensional data from the learned latent representation.
Our results demonstrate that the proposed methodology effectively captures complex data distributions while adhering to the underlying manifold structure, allowing for improved efficiency in generative sampling. Regarding the computational cost of each method, m-SGM1 and PLoM outperform m-SGM2, as the latter requires learning the SDE through a neural network. The ability to model high-dimensional densities constrained to lower-dimensional subspaces has significant implications for a wide range of applications, particularly in systems constrained by inherent geometric structures. The framework was validated through a benchmark problem and a multiscale material system, showcasing its robustness and potential in practical scenarios.
Future work will explore extensions of this framework to more complex and higher-dimensional datasets, as well as its integration with alternative manifold learning techniques. Furthermore, further refinement of the generative process, including adaptive sampling strategies and more efficient neural architectures, may enhance performance and applicability in diverse scientific and engineering domains.

\section*{Acknowledgments}
\noindent
Two of the authors (IGK and EC) are indebted to Dr. Juan Bello Rivas for many insightful discussions and suggestions on the subject of the paper. The authors are indebted to Dr. N. Evangelou for the use of his Double Diffusion Maps/GH code. The research efforts of the authors (DG and IGK) were supported by the U.S. Department of Energy (DOE) under Grant No. DE-SC0024162. 

\bibliographystyle{unsrt}  
\bibliography{manuscript}

\section*{APPENDIX}
\appendix
\section{Estimation of the Gradient of the Conditional Distribution} \label{sec:proof}

\noindent
The expression for the Gaussian conditional probability density is:
\[
f(x_t \mid x_0) = \kappa \exp\left(-\frac{1}{2\beta_t^2} \|x_t - \alpha_t x_0\|^2\right),
\]
where \( \|x_t - \alpha_t x_0\|^2 = (x_t - \alpha_t x_0)^\top (x_t - \alpha_t x_0) \) and $\kappa$ is a normalization constant:
\[
\kappa = \frac{1}{(2\pi)^{d/2} \beta_t^d}
\]
Since \( \kappa \) does not depend on \( x_t \), its gradient with respect to \( x_t \) is zero:
\[
\nabla_{x_t} \kappa = 0.
\]
\noindent
The gradient of the exponential term is:
\begin{equation}\label{eq:1}
\nabla_{x_t} \exp\left(-\frac{1}{2\beta_t^2} \|x_t - \alpha_t x_0\|^2\right) =
\exp\left(-\frac{1}{2\beta_t^2} \|x_t - \alpha_t x_0\|^2\right) \cdot \nabla_{x_t} \left(-\frac{1}{2\beta_t^2} \|x_t - \alpha_t x_0\|^2\right)
\end{equation}
where
\[
\nabla_{x_t} \left(-\frac{1}{2\beta_t^2} \|x_t - \alpha_t x_0\|^2\right) =
-\frac{1}{2\beta_t^2} \cdot \nabla_{x_t} \|x_t - \alpha_t x_0\|^2 = -\frac{1}{\beta_t^2} (x_t - \alpha_t x_0)
\]
Substituting back into the gradient of the exponential term in Eq.(\ref{eq:1}) we get:
\[
\nabla_{x_t} \exp\left(-\frac{1}{2\beta_t^2} \|x_t - \alpha_t x_0\|^2\right) =
\exp\left(-\frac{1}{2\beta_t^2} \|x_t - \alpha_t x_0\|^2\right) \cdot \left(-\frac{1}{\beta_t^2} (x_t - \alpha_t x_0)\right).
\]
Finally, multiplying this by the normalization constant \( \kappa \) we compute:
\[
\nabla_{x_t} f(x_t \mid x_0) = f(x_t \mid x_0) \cdot \left(-\frac{1}{\beta_t^2} (x_t - \alpha_t x_0)\right).
\]

The gradient of \( f(x_t \mid x_0) \) is:
\[
\nabla_{x_t} f(x_t \mid x_0) = -\frac{1}{\beta_t^2} (x_t - \alpha_t x_0) f(x_t \mid x_0).
\]

\section{Diffusion Maps}\label{sec:DMaps}

\subsection{Diffusion Maps for Dimensionality reduction}
\label{sec:dmaps_red}
\noindent
The Diffusion Maps algorithm \cite{coifman2006diffusion} can reveal the intrinsic geometry of the dataset \( \mathcal{X}=\{\textbf{x}_1, \textbf{x}_2, \ldots, \textbf{x}_N\} \) through the local similarity between sampled data points on \(\mathcal{X} \), i.e.,  a weighted graph \( \textbf{K} \in \mathbb{R}^{N \times N} \) based on a kernel function is constructed between data points. In this work, the following diffusion kernel was utilized:

\begin{equation}
    K(\textbf{x}_i, \textbf{x}_j) = \exp \left( -\frac{\| \textbf{x}_i - \textbf{x}_j \|^2}{2 \epsilon} \right),
\end{equation}
where \( \epsilon \) is a hyper-parameter that specifies the (square of the) kernel bandwidth \cite{gear2016manifolds}. Points \( \textbf{x}_i, \textbf{x}_j \) that satisfy \( \| \textbf{x}_i - \textbf{x}_j \|^2 < \epsilon \) are considered similar (connected), whereas points with a distance larger than \( \epsilon \) are effectively not linked. Different metrics may be used instead of the Euclidean distance. Diffusion Maps approximates the heat kernel \( e^{-t \Delta} \), where \( \Delta \) is the Laplace-Beltrami operator on a manifold \( \mathcal{M} \), in the limit as \( n \to \infty \) and \( \epsilon \to 0 \). Due to the influence of sampling density, normalization is needed to obtain the correct operator:

\begin{equation}
    \tilde{\textbf{K}} = \textbf{P}^{-\alpha} \textbf{K} \textbf{P}^{-\alpha},
\end{equation}
where $\mathbf{P}$ is a diagonal matrix with elements $P_{ii} = \sum_{j=1}^n K_{ij}$ and \( \alpha \) is a bias parameter controlling the sampling density's effect on geometry. Setting \( \alpha = 0 \) maximizes the density influence, while \( \alpha = 1 \) factors out density effects, providing a Laplace-Beltrami approximation. A second normalization of $\mathbf{\Tilde{K}}$ transforms it into a row-stochastic, Markovian matrix: $\mathbf{M} = \mathbf{D}^{-1} \mathbf{\Tilde{K}}$, where $\mathbf{D}$ is a diagonal matrix with elements $D_{ii} = \sum_{j=1}^n \Tilde{K}_{ij}$, yielding a row-stochastic, Markovian matrix. The eigendecomposition of \( \textbf{M{}} \) provides eigenvectors \( \phi_i \) and eigenvalues \( \lambda_i \):

\begin{equation}
    \textbf{M} \boldsymbol{\phi}_i = \lambda_i \boldsymbol{\phi}_i.
\end{equation}
Choosing independent eigenvectors \( \boldsymbol{\phi}_i \) for unique eigen-directions yields a parsimonious representation of the data \cite{dsilva2018parsimonious}. However, the selection of the eigenvectors should not be based solely on the eigenvalues, even though in some cases an \textit{eigengap} can provide a hint. This is because some eigenvectors parameterize the same direction. These eigenvectors are referred as harmonic eigenvectors and those that parameterize different directions as non-harmonic. The minimal set of eigenvectors should be restricted to selecting the non-harmonic eigenvectors. Specifically if \( k \), the number of independent non-harmonic eigenvectors, is smaller than the ambient space's dimension, \( \boldsymbol{\Phi} = \{ \boldsymbol{\phi}_{l} \} \), where $l = \{i_1, i_2, \dots, i_k\}, \quad i_j \in \mathbb{Z}_{+} \text{ for } j = 1, 2, \dots, N$, provides a lower-dimensional parametrization of the original dataset. 

\subsection{Diffusion Maps for lifting (Double Diffusion Maps)}
\label{sec:ddmaps}

\noindent
The Double Diffusion Maps scheme \cite{evangelou2023double} leverages Diffusion Maps for both dimensionality reduction and function extension. This method operates by obtaining a few reduced non-harmonic coordinates $\boldsymbol{\Phi}$  and constructing a basis $\boldsymbol{\Psi}$, called \textit{Latent Harmonics}, by operating solely on the principal non-harmonic eigenvectors. 
This basis serves as a global interpolation scheme, enabling scientific computations within this reduced latent space.  Note that, while  keeping only the primary non-harmonic eigenvectors and discarding the remaining eigenvectors (since they are dependent one the selected ones) reduces dimensionality, these discarded coordinates are essential for extending functions on the data manifold using GH. To restore the ability to extend functions while leveraging the intrinsic low dimensionality of the manifold, one need to reconstruct these harmonics. This is accomplished  by computing Latent Harmonics $\boldsymbol{\Psi}$ based on $\boldsymbol{\Phi}$. 

For each coordinate in the set $\{\boldsymbol{\phi}_i\}_{i=1}^N \in \Phi$, we compute $$\textbf{K}^*(\boldsymbol{\phi}_i, \boldsymbol{\phi}_j) = \exp\left(-\frac{||\boldsymbol{\phi}_i - \boldsymbol{\phi}_j||^2}{2\epsilon_2}\right)$$and calculate the first $m$ eigenvectors $\boldsymbol{\Psi} = \{\boldsymbol{\psi}_0, \dots, \boldsymbol{\psi}_{m -1}\}$, where $\boldsymbol{\psi}_i \in \mathbb{R}^N$, with corresponding eigenvalues $\boldsymbol{\sigma} = \{\boldsymbol{\sigma}_0, \dots, \boldsymbol{\sigma}_{m-1}\}$. Given the function $h$ defined on the data, we project $h$ on these eigenvectors

\begin{equation}
f \rightarrow P_\delta h = \sum_{j=1}^m \langle h, \boldsymbol{\psi}_j \rangle \boldsymbol{\psi}_j
\label{eq:mapping}
\end{equation}
and we compute the GH functions for $\boldsymbol{\phi}_{\text{new}}$ as:
\begin{equation}
\boldsymbol{\Psi}_j(\boldsymbol{\phi}_{\text{new}}) = \sigma_j^{-1} \sum_{i=1}^m \textbf{K}^*(\boldsymbol{\phi}_{\text{new}}, \boldsymbol{\phi}_i) \psi_j(\boldsymbol{\phi}_i)
\label{eq:psi_new}
\end{equation}
where $\psi_j(\boldsymbol{\phi}_i)$ is the $i$-th component of the $j$-th eigenvector.
Finally, we estimate the value of the function for $\boldsymbol{\phi}_{\text{new}}$ as:
    \begin{equation}
    (Ef)(\boldsymbol{\phi}_{\text{new}}) = \sum_{j=1}^m \langle h, \boldsymbol{\psi}_j \rangle \boldsymbol{\Psi}_j(\boldsymbol{\phi}_{\text{new}})
    \label{eq:func}
    \end{equation}

Algorithm \ref{alg:dim_reduction} depicts the basic steps for performing Double Diffusion Maps.

\begin{figure}[!htb]
    \centering
    \begin{subfigure}[b]{0.49\textwidth}
        \centering
        \includegraphics[width=\textwidth]{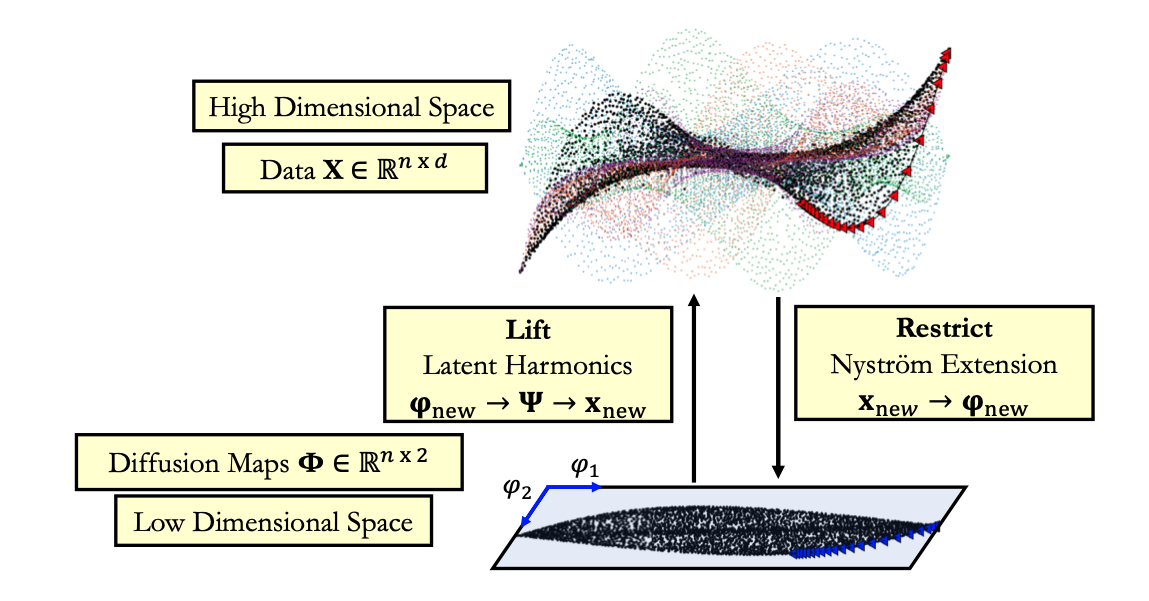}
        \caption{}
        \label{fig:sub1}
    \end{subfigure}
    \hfill
    \begin{subfigure}[b]{0.49\textwidth}
        \centering
        \includegraphics[width=\textwidth]{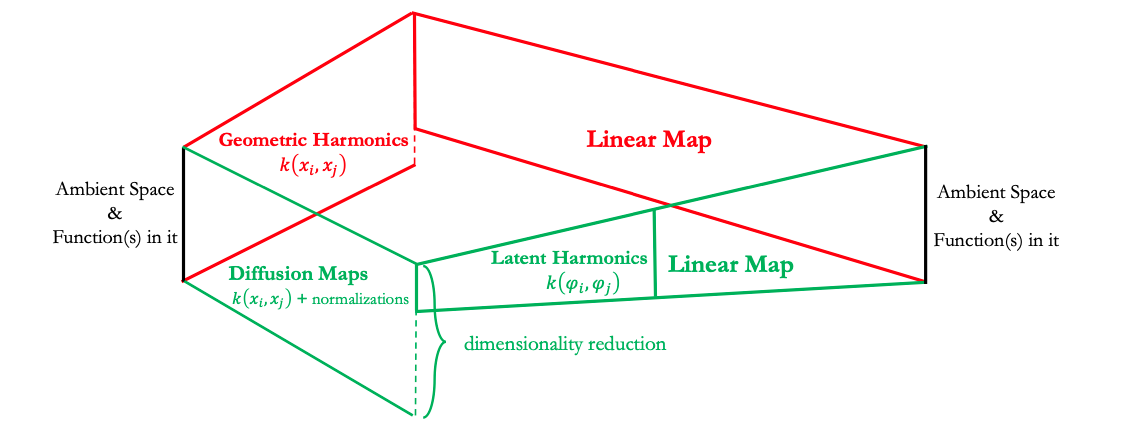}
        \caption{}
        \label{fig:sub2}
    \end{subfigure}
    \caption{(a) Double Diffusion Maps framework based on Diffusion Maps, Latent Harmonics, and Nystr\"om extension. (b) The upper red schematic illustrates mapping from the ambient space and its functions back to the same space using Geometric Harmonics. The lower green schematic shows the double Diffusion Maps approach: (i) using Diffusion Maps to uncover the intrinsic geometry of the dataset (potentially reducing dimensionality), and (ii) mapping from the resulting latent space back to the ambient space and its functions. Figures adopted from \cite{evangelou2023double}.}
    \label{fig:main}
\end{figure}

\begin{algorithm}[H]
\caption{Double Diffusion Maps}\label{alg:dim_reduction}

\begin{algorithmic}[1]
\Require A dataset $\mathcal{X}=\{\textbf{x}_i\}_{i=1}^N \subset \mathbb{R}^d \sim f_{\textbf{x}(\cdot)}$
\State \textbf{Diffusion Maps} for Dimensionality Reduction: 
\begin{itemize}
\item For each $\textbf{x}_i \in \mathcal{X}$, compute $K(\textbf{x}_i, \textbf{x}_j)$
\item Compute transition matrix $\tilde{\mathbf{K}} = \mathbf{P}^{-\alpha} \mathbf{K} \mathbf{P}^{-\alpha}$, where $P_{ii} = \sum_{j=1}^N K_{ij}$  and $\alpha$=1.

\item Normalize $\tilde{\mathbf{K}}$ and solve $\mathbf{M}\phi_i = \lambda_i \phi_i$ for each $i$.
\item Select the set of non-harmonic eigenvectors $\boldsymbol{\Phi} \in \mathbb{R}^{N \times k}$, where $k \leq m$.
\end{itemize}

\State \textbf{Double Diffusion Maps}:
\begin{itemize}
\item For each $\boldsymbol{\phi}_i \in \boldsymbol{\Phi}$, compute  $K^\star(\boldsymbol{\phi}_i, \boldsymbol{\phi}_j)$.
\item  Compute the $l$ first eigenvectors $\boldsymbol{\Psi} = \{ \boldsymbol{\psi}_0, \ldots, \boldsymbol{\psi}_{l-1} \}$ of $K^\star$, where $\boldsymbol{\psi}_i \in \mathbb{R}^N$.
\item Learn the mapping: $\mathsf{f} \to P_\delta \mathsf{f}$ using Eq.~(\ref{eq:mapping}).
\end{itemize}

\State \textbf{Latent Harmonics}:
\begin{itemize}
\State Compute the GH functions for $\boldsymbol{\Psi}_j(\boldsymbol{\phi}_\text{new})$  using Eq.~(\ref{eq:psi_new})
\item Estimate the value  $(E\mathsf{f})(\boldsymbol{\phi}_\text{new})$ with Eq.~(\ref{eq:func}).
\end{itemize}
\end{algorithmic}
\end{algorithm}

\section{Geometric Harmonics for function extension}\label{sec:GH}

\noindent
Geometric Harmonics is a technique inspired by the Nystr\"om method for extending empirical functions $h$ defined on a set $\mathcal{X}$ to a larger set $\mathcal{\overline{X}}$ \cite{fowlkes2001efficient}. 
In our case, GH can be used to generate a Diffusion Map coordinate $\boldsymbol{\phi}_{\text{new}}$ for a new sample point $\textbf{x}_{\text{new}} \notin \mathcal{X}$.  The mapping $\textbf{x}_{\text{new}} \rightarrow \boldsymbol{\phi}_{\text{new}}$ is called \textit{Restriction} and can be obtained by evaluating the kernel that was used during the dimensionality reduction step (and applying the same normalizations) on the new point $\textbf{x}_{\text{new}}$. More precisely, the Euclidean distance between this new point, $\textbf{x}_{\text{new}}$, and all the pre-existing points in the dataset $\mathcal{X}$ is computed as:

\begin{equation}
    \textbf{K}(\textbf{x}_{\text{new}}, \textbf{x}_j) = \exp\left(-\frac{||\textbf{x}_{\text{new}} - \textbf{x}_j||^2}{2\epsilon}\right),
\end{equation}

\begin{equation}
    \tilde{\textbf{K}}(\textbf{x}_{\text{new}}, \textbf{x}_j) = \frac{\textbf{K}(\textbf{x}_{\text{new}}, \textbf{x}_j)}{p(\textbf{x}_{\text{new}})^\alpha p(\textbf{x}_j)^\alpha},
\end{equation}
where $p(\textbf{x}_{\text{new}}) = \sum_{j=1}^N \textbf{K}(\textbf{x}_{\text{new}}, \textbf{x}_j)$ is a scalar value and $p(\textbf{x}_j) = \sum_{k=1}^N \textbf{K}(\textbf{x}_j, \textbf{x}_k)$ is an $N$-dimensional vector. Then, the matrix $\textbf{A}$ is constructed:

\begin{equation}
    \textbf{A}(\textbf{x}_{\text{new}}, \textbf{x}_i) = \frac{\tilde{\textbf{K}}(\textbf{x}_{\text{new}}, \textbf{x}_j)}{\sum_{j=1}^N \tilde{\textbf{K}}(\textbf{x}_{\text{new}}, \textbf{x}_j)},
\end{equation}
and the value of the reduced coordinate $\beta^{\text{th}}$ is computed as

\begin{equation}
    \phi_\beta(\textbf{x}_{\text{new}}) = \frac{1}{\lambda_\beta} \sum_{i=1}^N \textbf{A}(\textbf{x}_{\text{new}}, \textbf{x}_i) \phi_\beta(\textbf{x}_i),
\end{equation}
where $\phi_\beta(\textbf{x}_i)$ is the $i$-th component of the $\beta$-th eigenvector $(\phi_\beta)$ and $\lambda_\beta$ is the $\beta$-th eigenvalue. 




\section{Probabilistic Learning on the manifold for sampling densities}\label{sec:plom}

\noindent
Probabilistic learning on manifolds (PLoM) is designed to generate new realizations of random vectors whose probability distribution is concentrated on an unknown low-dimensional manifold embedded in a higher-dimensional space.  Detailed exposition of the procedures described here can be found in \cite{soize2019entropy, soize2021probabilistic, soize2022probabilistic}.

As a first step of PLoM, the training dataset, defined by matrix $[\phi_d]$, is construed as a realization from a latent random matrix variate $[\boldsymbol{\Phi}]$, and is reduced via a principal component analysis (PCA). The eigenvalues and eigenvectors $(\mu, \boldsymbol{\psi})$, of the covariance matrix $[c]$ are used to define a new random matrix variable $\textbf{H}$ through the relationship:

\begin{equation}\label{eq:pca}
[\boldsymbol{\Phi}] =\overline{[\phi]} + [\psi] [\mu]^{1/2} [\textbf{H}]
\end{equation}
where $\overline{[\phi]}$ is an $n \times N$ matrix whose each column is equal to the sample mean in $\mathbb{R}^n$ obtained from the $N$ samples, $[\psi]$ is a $n \times \nu$ matrix whose columns contain the dominant eigenvectors of $[c]$, and $[\mu]$ is a $\nu \times \nu$ diagonal matrix with
the corresponding eigenvalues. The PCA reduction expressed by captures information in the original data $[\phi_d]$ as encoded in the linear correlation between the $n$ features, averaged over the $N$ data points. It relies for its fidelity on the
n-dimensional eigenvectors of the covariance matrix $[c]$.  

Letting  matrix $[\eta_d]$ denote the PCA coordinates of the training set $[\phi_d]$, an estimate of the probability measure of the $\mathbb{R}^{\nu}$-valued random vector $[\textbf{H}]$ can then be obtained in the form of the following Gaussian mixture,

\begin{equation}\label{eq:pdf}
f_{\textbf{H}}(\boldsymbol{\eta}) = \frac{1}{N} \sum_{j=1}^N \pi_{\nu,\hat{s}_\nu} \left( \frac{\hat{s}_\nu}{s_\nu} \boldsymbol{\eta}_d^j - \boldsymbol{\eta} \right)
\end{equation}
where $\boldsymbol{\eta}_d^j$ is the $j^{th}$ column of $[\boldmath{\eta}_d]$  and $\pi_{\nu,\hat{s}_\nu}(\boldsymbol{\eta})$ is the isotropic zero-mean Gaussian density function in $\mathbb{R}^\nu$ defined as:
\begin{equation}
\pi_{\nu,\hat{s}_\nu}(\boldsymbol{\eta}) = \frac{1}{(\sqrt{2\pi} \hat{s}_\nu)^\nu} \exp\left(-\frac{1}{2 \hat{s}_\nu^2} \|\boldsymbol{\eta}\|^2\right)
\end{equation}
where
\begin{equation}
s_\nu = \left\{ \frac{4}{N(2+\nu)} \right\}^{\frac{1}{\nu+4}}, \quad \hat{s}_\nu = s_\nu \sqrt{s_\nu^2 + \frac{N-1}{N}}
\end{equation}
are positive parameters. With this choice of $s_\nu$ and $\hat{s}_\nu$ the mean-squared error is minimized and realizations of random vector $[\textbf{H}]$ are normalized. Since the $N$ columns of $[\textbf{H}]$
as statistically independent, the density of $[\textbf{H}]$ is given by:

\begin{equation}
  f_{[\textbf{H}]}([\boldsymbol{\eta}])=  f_{[\textbf{H}]}(\boldsymbol{\eta}^1)\times \ldots \times f_{[\textbf{H}]}(\boldsymbol{\eta}^N)
\end{equation}
To generate new realizations, the sampling process is confined to the manifold by solving a projected  It\^o Stochastic Differential Equation (ISDE). This approach leverages a reduced-order representation of $[\textbf{H}]$ obtained with Diffusion Maps \cite{coifman2006diffusion} (see \ref{sec:dmaps_red}) to construct a stochastic process confined to the low-dimensional manifold. 

 The diffusion map embedding provides a parameterization
of points using in an $m$-dimensional subspace of $\mathbb{R}^N$, permitting us to express random matrix $[\mathbf{H}]$ as,

\begin{equation}
    [\mathbf{H}] = [\mathbf{Z}] [g]^\top,
\end{equation}
where $\mathbf{Z} \in \mathbb{R}^{\nu \times m}$ is a random matrix, and $[g] \in \mathbb{R}^{N\times m}$ is a matrix whose columns are the $m$ dominant Diffusion Map basis vectors. The reduced-order ISDE is formulated to generate independent samples of  $\mathbf{Z}$. Introducing $\mathbf{Z}(r)$ and $\mathbf{Y}(r)$ as the reduced-order stochastic processes for position and velocity, respectively, the ISDE is expressed as:
\begin{align}
    d[\mathbf{Z}(r)] &= \mathbf{Y}(r)dr, \label{eq:isde_pos} \\
    d[\mathbf{Y}(r)] &= [\mathbf{L}([\mathbf{Z}(r)][g]^\top)][\alpha]dr - \frac{f_0}{2} [Y(r)]dr + \sqrt{f_0} \, dW(r)dr. \label{eq:isde_vel}
\end{align}
with the initial conditions
\begin{equation}
    [\textbf{Z}(0)] = [\eta_d][\alpha], \quad [\textbf{Y}(0)] = [\mathcal{N}][\alpha]\ .
\end{equation}
Here, $[\alpha] = [g]([g]^\top [g])^{-1}$,
$[\mathcal{N}]\in \mathbb{R}^{\nu \times N}$  with entries that are independent standard Gaussian variables, $f_0$ is a dissipation parameter controlling the damping in the stochastic system, $dW(r)$ represents the increments of a Wiener process, and  $[\mathbf{L}([u])]_{kl}$ is a force term defined by the gradient of the potential function:
    \begin{equation}
        [\mathbf{L}([u])]_{kl} = \frac{\partial}{\partial u_k^l} \log \{q(\textbf{u}^l)\}, \quad \textbf{u}^l = (u_1^l, u_2^l \ldots, u_{\nu}^l), \quad [u] = [\textbf{u}^1, \ldots, \textbf{u}^N]
    \end{equation}
   where where $\textbf{u}^l \rightarrow q(\textbf{u}^l)$  is the continuously differentiable function such that 
   \begin{equation}\label{eq:eq}
q(\textbf{u}^l) = \frac{1}{N} \sum_{j=1}^N \exp{\{-\frac{1}{2s_{\nu}^2} } \left( \frac{\hat{s}_\nu}{s_\nu} \boldsymbol{\eta}_d^j - \textbf{u}^l \right)\}
\end{equation}

    In this work, the reduced-order ISDE is solved numerically using the Störmer–Verlet scheme, which is efficient for weakly dissipative Hamiltonian systems. 

\end{document}